%% file: arxiv_paassen_hammer_price_barnes_gross_pinkwart_2017_continuous_hint_factory.tex
\documentclass[a4,12pt]{article} 

\usepackage{amsmath}
\usepackage{amsthm}
\usepackage{amssymb}
\usepackage{mathtools}
\usepackage{bm}
\usepackage[svgnames]{xcolor}
\usepackage{pdfpages}
\usepackage{listings}
\usepackage{algorithm}
\usepackage{algpseudocode}
\usepackage{tango-colors}
\usepackage{tikz}
\usetikzlibrary{arrows}
\usetikzlibrary{shapes}
\usepackage{pgfplots}
\pgfplotsset{compat=1.9}
\usepgfplotslibrary{groupplots}
\usepackage{pgfplotstable}
\usepackage{multirow}
\usepackage{booktabs}

\usepackage{enumitem}
\usepackage{subcaption}

\usepackage[round,comma,authoryear]{natbib}
\newcommand{\citeN}[1]{\citet{#1}}
\renewcommand{\cite}[1]{\citep{#1}}
\usepackage{doi}

\usepackage[utf8]{inputenc}
\usepackage[T1]{fontenc}
\usepackage[english]{babel}
\usepackage{hyperref}
\usepackage{csquotes}

\input{symbols}

\newtheorem{thm}{Theorem}

\theoremstyle{definition}
\newtheorem{dfn}{Definition}

\makeatletter
\@addtoreset{hyp}{section}
\makeatother

\newcommand{\mail}[1]{\href{mailto:#1}{#1}}

\begin{document}

\title{The Continuous Hint Factory - Providing Hints in Vast and Sparsely Populated Edit Distance Spaces}

\author{%
\begin{tabular}[t]{ll}
{\large Benjamin Paaßen} & {\large Barbara Hammer} \\
CITEC center of excellence & CITEC center of excellence \\
\mail{bpaassen@techfak.uni-bielefeld.de} & \mail{bhammer@techfak.uni-bielefeld.de} \vspace{0.2cm}\\
{\large Thomas W.\ Price} & {\large Tiffany Barnes} \\
North Carolina State University & North Carolina State University \\
\mail{twprice@ncsu.edu} & \mail{tmbarnes@ncsu.edu}\vspace{0.2cm}\\
{\large Sebastian Gross} & {\large Niels Pinkwart} \\
Humboldt-Universität zu Berlin & Humboldt-Universität zu Berlin \\
\mail{sebastian.gross@informatik.hu-berlin.de} & \mail{niels.pinkwart@hu-berlin.de}
\end{tabular}
}

\date{This is a preprint of the publication \citeN{Paassen2018JEDM} as provided by the authors.}

\pagestyle{myheadings}
\markright{Preprint as provided by the authors.}

\maketitle

\input{content.tex}

\bibliographystyle{plainnat}
\bibliography{literature.bib}

\end{document}

%% file: symbols.tex
\newcommand{\effic}{\mathcal{O}}

\newcommand{\R}{\mathbb{R}}

\newcommand{\dims}{s}

\newcommand{\N}{\mathbb{N}}

\newcommand{\mat}[1]{\bm{#1}}

\newcommand{\transp}[1]{{ #1 }^T}
\newcommand{\eye}{I}

\newcommand{\graph}{G}
\newcommand{\nodes}{V}
\newcommand{\node}{v}
\newcommand{\lnode}{u}
\newcommand{\rnode}{\node}
\newcommand{\nodeidx}{n}
\newcommand{\nodelim}{N}

\newcommand{\edges}{E}

\newcommand{\elabel}{\beta}

\newcommand{\kernel}{k}
\newcommand{\kernelfeat}{\vec \phi}
\newcommand{\kernelspace}{\mathcal{Y}}
\newcommand{\kernelvec}{\vec{\mathrm{k}}}
\newcommand{\kernelmat}{\mat{K}}

\newcommand{\bandwidth}{\psi}

\newcommand{\prob}{P}

\newcommand{\nDev}{\sigma}
\newcommand{\nDevNoise}{\tilde \nDev}

\newcommand{\pol}{\pi}

\newcommand{\dataidx}{i}
\newcommand{\ldataidx}{\dataidx}
\newcommand{\rdataidx}{j}
\newcommand{\datalim}{M}

\newcommand{\outstate}{y}

\newcommand{\tracelim}{N}
\newcommand{\traceidx}{i}
\newcommand{\timelim}{T}
\newcommand{\timeidx}{t}

\newcommand{\states}{X}

\newcommand{\state}{x}
\newcommand{\lstate}{x}
\newcommand{\rstate}{y}

\newcommand{\dist}{d}
\newcommand{\distmat}{D}

\newcommand{\coeff}{\alpha}
\newcommand{\sparselim}{m}

\newcommand{\gpcoeff}{\gamma}

\newcommand{\editCost}{C}

\newcommand{\del}{\mathrm{del}}
\newcommand{\rep}{\mathrm{rep}}
\newcommand{\ins}{\mathrm{ins}}

\newcommand{\edit}{\delta}
\newcommand{\edits}{\Delta}

\newcommand{\editvec}{{\vec \xi}}

%% file: content.tex
\begin{abstract}
Intelligent tutoring systems can support students in solving multi-step tasks by providing
hints regarding what to do next. However, engineering such next-step hints manually or via an
expert model becomes infeasible if the space of possible states is too large.
Therefore, several approaches have emerged to infer next-step hints automatically, relying on past
students' data. In particular, the Hint Factory \cite{Barnes2008} recommends edits that are
most likely to guide students from their current state towards a correct solution, based on what
successful students in the past have done in the same situation. Still, the Hint Factory relies on
student data being available for any state a student might visit while solving the task, which is
not the case for some learning tasks, such as open-ended programming tasks.
In this contribution we provide a mathematical framework for edit-based hint policies
and, based on this theory, propose a novel hint policy to provide edit hints
in vast and sparsely populated state spaces. In particular, we extend the Hint Factory
by considering data of past students in all states which are similar to the student's current
state and creating hints approximating the weighted average of all these reference states.
Because the space of possible weighted averages is continuous, we call this approach the
Continuous Hint Factory.
In our experimental evaluation, we demonstrate that the Continuous Hint Factory can predict 
more accurately what capable students would do compared to existing prediction schemes on two
learning tasks, especially in an open-ended programming task, and that the Continuous Hint Factory
is comparable to existing hint policies at reproducing tutor hints on a simple UML diagram task.
\end{abstract}

\emph{keywords: next-step hints, Hint Factory, edit distances, computer science education, Gaussian Processes}

\section{Introduction}

In many educational domains, learning tasks require more than a single step to solve. For example,
programming tasks require a student to iteratively write, test, and refine code that accomplishes a
given objective \cite{Gross2014,Price2017,Rivers2015}. When working on such multi-step-tasks,
students start with an initial state and then change their state by applying an action (such as
inserting or deleting a piece of code). At some point, a student may not know how to proceed or may
be unable to find an error in her current state, in which case external help is required. In
particular, such a student may benefit from a next-step hint, guiding her toward a more complete
and/or more correct version and allowing her to continue working on her own \cite{Aleven2016}.
Many intelligent tutoring systems (ITSs) attempt to create such next-step hints automatically,
and adjust such hints to the student's current state as well as her underlying strategy 
\cite{VanLehn2006}. Typically, hints are created using an expert-crafted model. However, 
designing such an expert model becomes infeasible if the space of possible states is to variable
to cover with expert rules \cite{Murray2003,Koedinger2013,Rivers2015}. 
This is the case for most computer programming tasks because the space of possible programs grows 
exponentially with the program length and the set of programs which perform the same function is 
infinite \cite{Piech2015}. Other examples are so-called ill-defined domains where explicit domain 
knowledge is not available or at least very hard to formalize \cite{Lynch2009}.

Several approaches have emerged which provide next-step hints without an expert model. Typically, 
these approaches provide hints in the form of \emph{edits}, that is, actions which
can be applied to the student's current state to change it into a more correct and/or more
complete state, based on the edits that successful students in the past have applied
\cite{Gross2015AIED,Price2016,Rivers2015,Zimmerman2015}. Such edit-based
next-step hints constitute an elegant and simple approach to feedback for complex learning tasks.
The most basic version of the approach requires only two ingredients: a function, which is able
to compute the shortest sequence of edits to transform one partial solution to the 
task into another one, and a correct solution for the task. If a student issues a help request, the 
system can simply compute the edits from the student's state to the solution and use one 
of these edits as a hint \cite{Rivers2015,Zimmerman2015}. Even though this approach is fairly 
simple, it achieves \emph{personalized} feedback, in the sense that the hint
depends on the student's personal state and may thus be fitting to her specific strategy and style 
\cite{Le2014}. Further, this approach needs very little task-specific work on the side of domain 
experts because they only need to construct example solutions for the task and can
apply a general-purpose edit function which is applicable across tasks or even across domains
\cite{Mokbel2013EDM}.

A key challenge to such an edit-based hint approach is that it attempts to follow a single reference
solution and tries to adjust the student's individual solution in all aspects to the reference
solution. In contrast, it may be more desirable to provide hints which correspond to what capable
students would do in any given situation \emph{in general} but avoid emulating specific style choices
by individual solutions.
To identify such generic solution steps, most existing approaches rely on
\emph{frequency} information, that is, how often a certain edit or state has occurred in past
student's data \cite{Barnes2008,Lazar2014,Rivers2014,Piech2015}.
Unfortunately, for many programming tasks, the space of possible programs is so large that hardly 
any state is visited more than once, even if aggressive pre-processing methods are applied to 
canonicalize program representations \cite{Price2015}.

Therefore, a novel approach is needed which can select generic edits even in cases where 
frequency information is not available. We base this approach on the Hint Factory, which generates
hints that have led past students in the same situation to a correct solution
\cite{Barnes2008,Stamper2012}. To transfer this approach to vast and sparsely populated spaces,
we consider not only the data of past students who have visited the same state, but also
\emph{similar} states, and we represent the reference state to which a student should move as a
weighted average of past students' states, where the weights are chosen in a probabilistically
optimal sense. Because the space of possible weighted averages is continuous, our reference states
exist in an implicit, continuous state, which is why we call our approach the 
\emph{Continuous Hint Factory} (CHF). By performing a weighted average, we avoid any individual
specificities and focus on generic steps toward a correct solution.

More precisely, the key contributions of our paper are as follows: First, we provide precise 
definitions of key concepts in the field of edit-based hint policies and integrate them into 
a mathematical framework. Second, we extend this framework by introducing the notion of an 
\emph{edit distance space}, a continuous space in which each state corresponds to one vector and the 
Euclidean distance between vectors corresponds to the edit distance between two states. Finally, 
we utilize this space for the CHF by identifying the most likely hint as a vector in this space and 
translating this vector back into a human-readable edit.
As such, the CHF constitutes a novel hint technique which is applicable  whenever an edit distance
and some, possibly few, data samples of past students are available, making it an interesting option
for vast and sparsely populated state spaces. For example, such spaces occur in programming tasks
where a solution involves many lines of codes and solutions are highly variable in terms of
strategy and style.

In experiments on two datasets we provide evidence that the CHF is 
able to predict what capable students would do in solving a learning task, that the CHF is able to
disambiguate between many possible edits, and that the hints provided by the CHF match the hints of
human tutors at least as well as other established hint techniques.

We begin our work by introducing precise definitions of key concepts of edit-based hint policies
and review existing approaches within this novel framework. In Section~\ref{sec:chf} we introduce
the Continuous Hint Factory based on this framework and in Section~\ref{sec:experiments} we report
on our experimental evaluation of the Continuous Hint Factory.

\section{An integrated view of edit-based hint policies}

\begin{figure}
\begin{center}
\begin{subfigure}[t]{0.37\textwidth}
\centering
\includegraphics[width=\textwidth]{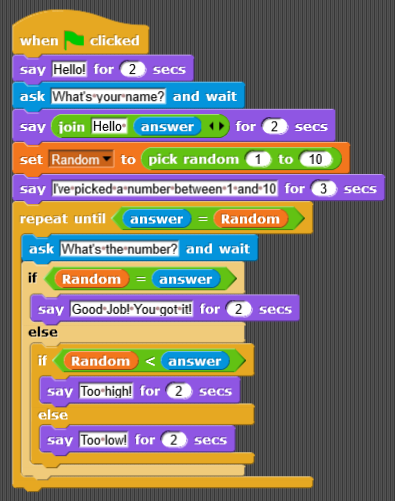}
\caption{}
\label{fig:snap}
\end{subfigure}
\hspace{0.03\textwidth}
\begin{subfigure}[t]{0.55\textwidth}
\centering
\begin{tikzpicture}
\begin{axis}[
	legend pos=south east,
	cycle list name=tangolist,
	width=\textwidth,
]
\foreach \t in {1,2,...,10} {
\addplot+[
quiver={
	u=\thisrow{deltax},
	v=\thisrow{deltay}
},
-stealth,
semithick]
table[x=x, y=y]
{results_data/snap_trajectory\t.csv};
}
\end{axis}
\end{tikzpicture}
\caption{}
\label{fig:snap_embedding}
\end{subfigure}
\end{center}
\caption{(a) A screenshot from the Snap programming environment. (b) A 2D embedding of ten 
example traces in the Snap dataset The 2D embedding was obtained via non-metric multi-dimensional
scaling \protect\cite{Sammon1969} using the pairwise edit distances as input, that is, if two states have an
edit distance of $0$, they are mapped to the same point in 2D, and if they have a higher edit
distance, they are mapped to points which are further apart. Colors are used to distinguish between
different traces. States within one trace are connected by arrows.}
\end{figure}
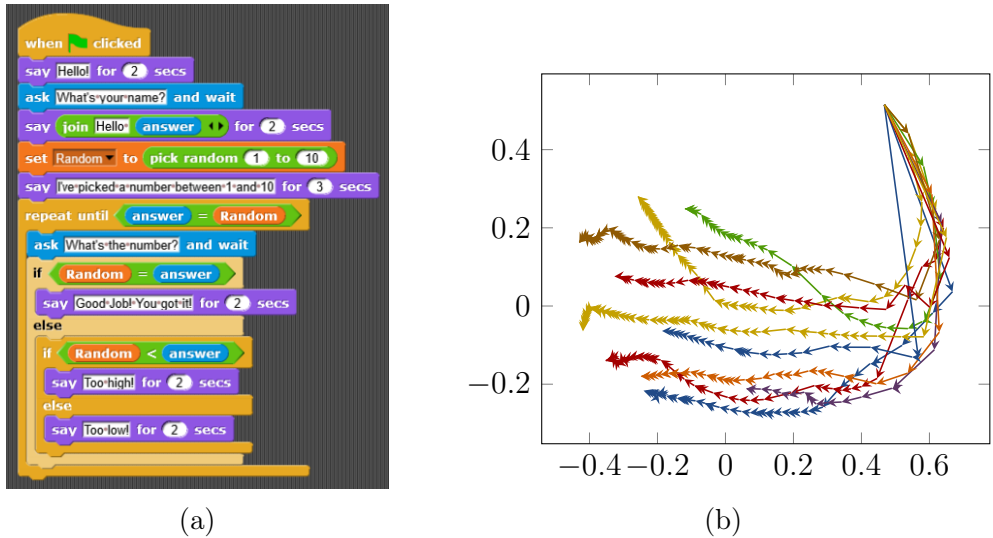

In this section, we review existing approaches to edit-based hint policies. Alongside this review,
we develop a mathematical theory of edit distances and edit-based hint policies, which helps us to
contextualize past approaches and will form a firm basis for our own approach in the next section.
First, let us start with an example of a learning task, for which an edit-based hint policy may be
helpful. Consider the task of programming a guessing game, which should ask the 
player for their name, generate a random number between 1 and 10, and then let the player guess the 
number, providing feedback to the player regarding whether the number was too low, too high, or
correct. A correct solution for this task in the \emph{Snap} programming 
language\footnote{\url{http://snap.berkeley.edu}} is shown in Figure~\ref{fig:snap}. In a tutoring
system involving this task, a student would start off with an empty program and then would add 
blocks to the program, delete them, or replace them with others until the student obtains a correct
solution or gets stuck. In the latter case, the student may hit a \enquote{help} button in the
system which would, in turn, provide a hint in the form of an edit which leads the student closer to a
correct solution (e.g., to add an \enquote{ask} block to ask for the player's name in the beginning).

From a pedagogical point of view, it may be suboptimal to immediately tell the student which edit 
to apply. In doing so, we deprive students of the possibility of finding the correct next step 
themselves and do not require the students to reflect on underlying concepts, as suggested by 
\citeN{Fleming1993} as well as \citeN{Le2016}. Indeed, \citeN{Aleven2016} suggest displaying such
hints, which reveal the next step, only as a last resort after 
exhausting options for more principle-based hints. This begs the question why the
focus of our work lies on such bottom-out hints.

First, edit hints are different from other bottom-out hints in that they display only a very small
part of the solution (a single edit), allowing the student to finish most of the problem
themselves. Second, bottom-out hints may lead to learning gains if students
reflect on the hint and engage in sense-making behavior \cite{Aleven2016,Shih2008}. Conversely, if 
students aim to abuse the system, this is not hindered by principle-based hints: students simply 
skip through such hints to reach the bottom-out hint \cite{Aleven2016,Shih2008}.
Third, we point to a study by \citeN{Price2017AIED} which indicates that edit hints are judged
as relevant and interpretable by human tutors.
Finally, and most importantly, we argue that more elaborate hint strategies are simply not
available in many important learning tasks because they require expert-crafted hint messages 
which are difficult to apply at scale \cite{Le2014,Murray2003,Rivers2015}.

In particular, there have been some approaches to make expert-crafted hints available in
larger state spaces, for example, authoring tools for tutoring systems, which aim at
reducing the expert work that needs to be put in to design feedback for individual tasks.
A prime example are the Cognitive Tutor Authoring Tools (CTAT), which support the construction of
cognitive tutors \cite{Aleven2006}. Cognitive tutors can be seen as a gold standard of intelligent 
tutoring systems because their effectiveness has been established in classroom studies, and they
have been successfully applied in classrooms across the US \cite{Koedinger2013,Pane2014}.
However, even with authoring tools, covering all possible variations in a suficiently variable
state space with many viable solutions may be infeasible \cite{Le2014,Murray2003,Rivers2015}.
For example, in our programming dataset (see Figure~\ref{fig:snap}), we consider more than
40 different solution strategies, each of which involves more than 40 steps.

Another approach is \enquote{force multiplication,} which assumes that a relatively small number
of expert-crafted hint messages are available, which are then applied to new situations
automatically, thereby \enquote{multiplying the force} of expert work \cite{Piech2015ICML}.
Examples include the work of \citeN{Choudhury2016}, \citeN{Head2017}, as well as \citeN{Yin2015} who apply
clustering methods to aggregate many different states and then provide the same hint to all
states in the same cluster.
Another example is the work of \citeN{Piech2015ICML} who let experts develop hints which are annotated
with example states for which the respective hint makes sense and example states for which
the respective hint does \emph{not} make sense.
Then, they train a classifier function for each hint via machine learning which decides for any new
state whether the hint should be displayed or not.
Finally, \citeN{Marin2017} annotate expert-crafted hints with small snippets of Java code for which
the given hint makes sense and then display the hint whenever the respective snippet is discovered
in a student's state.
Note that these approaches are limited by the number of hints that are provided by the teaching
experts. If for some situation no hint is contained in the database, the system is not able
to provide fitting feedback.

Due to these scaling limitations, we focus on edit-based bottom-out hints, which are easy to
individualize and generate automatically, as ample work in the past has demonstrated
\cite{Gross2015AIED,Lazar2014,Price2016,Rivers2015,Zimmerman2015}.

In the remainder of this section, we will analyze edit-based next-step hint approaches in more
detail. We will highlight key concepts, provide precise definitions and shed some light
on the theory behind edit-based hint approaches.
We start our investigation by defining edits, legal move graphs, and edit distances in a rigorous
fashion. Second, we discuss techniques to change our data representation in order to support
meaningful hints. Third, we incorporate student data in the form of traces and interaction networks.
Finally, we provide an overview of the different approaches that have emerged in the literature
and compare them in light of our mathematical framework.

\subsection{Edit Distances and Legal Move Graphs}

Recall that we wish to support students in solving a multi-step learning task by providing on-demand
edit hints. More precisely, we assume the following scenario. A student starts in some initial
state provided by the system, and then successively edits this initial state until she finishes the
task or gets stuck and asks the system for help. An edit hint for such a case should be an edit
which gets  the student closer to a desirable next state. To formalize this notion, we first define
what kind  of edits are possible at all (the \emph{edit set}). Then, we define how to combine such
edits to transform some student state into another (the \emph{legal move graph}). Finally, we can
define a  notion of \emph{distance} between states based on how many edits are necessary at least to 
transform one state into another (the \emph{edit distance}).

The notion of an edit set should cover all actions which a student can perform to change 
their current partial solution to a different state. Recall our example of the guessing game 
programming task in Figure~\ref{fig:snap}. In this scenario, the set of possible states is the 
set of possible Snap programs.
The possible edits are to add a block at any point in the 
program, replacing a block with another one, or deleting a block. For example, we may delete the 
\enquote{say 'Hello!' for 2 secs} block in Figure~\ref{fig:snap} or replace it with a \enquote{say 
'Hello!' for 1 sec}-block. Note that this edit set is \emph{symmetric}, in the sense that we can 
reverse every edit we have applied by deleting an inserted block, re-inserting a deleted block, or 
replacing a replaced block with its prior version.
This is a desirable property for edit sets because it ensures that any action of a student can be 
reversed and thus any state that can be reached by edits from the initial state is reachable from 
each other state. Formally, we define edit sets as follows.

\begin{dfn}[Edit Set]
Let $\states$ be the set of all possible states for a learning task. Then we call $\states$ the
\emph{state space} of the learning task.
An \emph{edit} on $\states$ is a function $\edit : \states \to \states$.
A set $\edits$ of edits on $\states$ is called an \emph{edit set}. 
We call an edit set \emph{symmetric} if for all states $\lstate \in \states$ and all edits $\edit 
\in \edits$ there exists an edit $\edit^{-1} \in \edits$ such that $\edit^{-1}(\edit(\lstate)) = 
\lstate$. We call $\edit^{-1}$ an \emph{inverse edit} for $\edit$ on $\lstate$.
\end{dfn}

Formally, our goal is to devise a function which can, for any state students may find themselves 
in, return an edit they should apply. Inspired by \citeN{Piech2015} we call such a function a 
\emph{hint policy}.\footnote{Note that \citeN{Piech2015} define a hint policy 
differently, namely as a function $\pol'$ mapping a state to a state the student should proceed to
next. Our definition presented here is a proper generalization of Piech et al.'s definition because
every policy $\pol$ according to our definition we can be converted into a Piech-style hint policy
$\pol'$ by setting $\pol'(\state) = \edit(\state)$ where $\edit = \pol(\state)$.}

\begin{dfn}[Hint Policy] \label{dfn:hint_policy}
Let $\states$ be a set and $\edits$ be an edit set on $\states$. A \emph{hint policy} is a function
$\pol : \states \to \edits$.
\end{dfn}

A helpful hint policy should return a hint that gets students closer to a desirable next state. For 
example, the hint policy of \citeN{Zimmerman2015} recommends the first edit in the shortest  
sequence of edits which transforms the student's current state to a correct
solution for the task. To develop such a hint policy, we need to properly 
define what \emph{closeness} between two states means and what the shortest sequence of edits is.
To provide an intuitive meaning to closeness and shortest sequences, we can rely on graph theory.
In particular, we can regard the state space as nodes of a directed graph and the edits as edges
in that graph. More precisely, we draw an edge from a state $\lstate$ to another state $\rstate$ if
and only if there is an edit in the edit sets $\edits$ which transforms $\lstate$ into $\rstate$.
This results in the notion of a \emph{legal move graph} \cite{Piech2015}.

\begin{dfn}[Legal Move Graph]
Let $\states$ be a set and $\edits$ be an edit set on $\states$. Then, the legal move
graph according to $\states$ and $\edits$ is defined as the directed graph
$\graph_{\states, \edits} = (\states, \edges)$ where
$\edges = \{ (\lstate, \rstate) | \exists \edit \in \edits : \edit(\lstate) = \rstate \}$.
\end{dfn}

For our programming example in Figure~\ref{fig:snap}, the legal move graph is too large to list 
here. Instead, consider the set of strings $\states = \{ \text{a}, \text{aa}, \text{aac}, \text{ab},
\text{abc}, \text{b}, \text{bb}, \text{bbc} \}$ and as edit set $\edits$ consider deletions, 
replacements, and insertions of single characters in these strings. More precisely, we have 
\begin{align}
\edits &= \big\{ \del_\nodeidx, \ins_{\nodeidx, \lnode}, \rep_{\nodeidx, \lnode} | \nodeidx \in \N, 
\lnode \in \{ \text{a}, \text{b}, \text{c}\}  \big\} & \label{eq:string_edit_set} \text{where} \\
\del_\nodeidx(\rnode_1, \ldots, \rnode_\nodelim) &= \rnode_1, \ldots, \rnode_{\nodeidx-1}, 
\rnode_{\nodeidx+1}, \ldots, \rnode_\nodelim \\
\ins_{\nodeidx, \lnode}(\rnode_1, \ldots, \rnode_\nodelim) &= \rnode_1, \ldots, \rnode_\nodeidx, 
\lnode, \rnode_{\nodeidx+1}, \ldots, \rnode_\nodelim \\
\rep_{\nodeidx, \lnode}(\rnode_1, \ldots, \rnode_\nodelim) &= \rnode_1, \ldots, \rnode_{\nodeidx-1}, 
\lnode, \rnode_{\nodeidx+1}, \ldots, \rnode_\nodelim
\end{align}
An excerpt of the legal move graph for this example is shown in
Figure~\ref{fig:chf_legal_move_graph}. In particular, \enquote{ab} is connected to \enquote{a},
\enquote{aa}, \enquote{b}, \enquote{bb}, and \enquote{abc}
because we can delete b, replace b with a, delete a, replace a with b, and insert c 
to transform \enquote{ab} to the respective other strings. Note that all arrows in this legal move 
graph are bi-directional, indicating the symmetry of the edit set.

\tikzstyle{state}=[circle, draw=aluminium6, fill=aluminium3, thick]
\tikzstyle{defaultcolor}=[aluminium6]
\tikzstyle{edge}=[->,shorten >=1pt,shorten <=1pt, draw=aluminium6, fill=aluminium6, thick, 
>=stealth']

\tikzstyle{trace}=[edge]
\tikzstyle{user1color}=[scarletred3]
\tikzstyle{user1}=[draw=scarletred3, fill=scarletred1, text=scarletred3]
\tikzstyle{user2color}=[skyblue3]
\tikzstyle{user2}=[draw=skyblue3, fill=skyblue1, text=skyblue3]
\tikzstyle{user3color}=[orange3]
\tikzstyle{user3}=[draw=orange3, fill=orange1, text=orange3]
\tikzstyle{user4color}=[plum3]
\tikzstyle{user4}=[draw=plum3, fill=plum1, text=plum3]

\tikzstyle{semclass}=[line width=0.02cm, rounded corners, draw=aluminium6, fill=aluminium1]
\tikzstyle{hint1}=[draw=orange3,fill=orange1,text=orange3]
\tikzstyle{hint1color}=[orange3]
\tikzstyle{hint2}=[draw=plum3,fill=plum1]
\tikzstyle{hint2color}=[plum3]
\tikzstyle{hint3}=[draw=chameleon3,fill=chameleon1]
\tikzstyle{hint3color}=[chameleon3]
\tikzstyle{hint4}=[draw=butter3,fill=butter1]
\tikzstyle{hint4color}=[butter3]

\begin{figure}
\begin{subfigure}[t]{0.3\textwidth}
\begin{center}
\begin{tikzpicture}[scale=1.5]
\begin{scope}[shift={(-1, 0)}]
\node[user2color]   (x1) [label={[user2color]above:$\state_1$}]    at (140:0.9) {a};
\node[user2color]   (x2) [label={[user2color]below:$\state_2$}]    at (220:0.9) {b};
\end{scope}

\node[user1color]   (x)  [label={[user1color]left:$\state$}]       at (-1, 0) {ab};

\node[defaultcolor] (aa)                                           at (120:1) {aa};
\node[user2color]   (y1) [label={[user2color]above:$\outstate_1$}] at ( 60:1) {aac};

\node[defaultcolor] (bb)                                           at (240:1) {bb};
\node[user2color]   (y2) [label={[user2color]below:$\outstate_2$}] at (300:1) {bbc};

\node[defaultcolor] (h) at (1, 0) {abc};

\path[edge, user2]%
(x1) edge [bend left]  (y1)
(x2) edge [bend right] (y2);

\path[edge, <->]%
(x)  edge (x1) edge (x2) edge (aa) edge (bb) edge (h)
(x1) edge (aa) edge (x2)
(aa) edge (y1)
(y1) edge (h)
(x2) edge (bb)
(bb) edge (y2)
(y2) edge (h);

\end{tikzpicture}
\end{center}
\caption{The legal move graph using the edit set of the string edit distance on the state space
$\states = \{ \text{a}, \text{aa}, \text{aac}, \text{ab},
\text{abc}, \text{b}, \text{bb}, \text{bbc} \}$.
$\state = \text{ab}$ is the current student state (red). Further, two traces are given with the 
states $\state_1 = \text{a}, \outstate_1 = \text{aac}$, and $\state_2 = \text{b}, \outstate_2 = 
\text{bbc}$ respectively (blue).}
\label{fig:chf_legal_move_graph}
\end{subfigure}
\hspace{0.03\textwidth}
\begin{subfigure}[t]{0.3\textwidth}
\begin{center}
\begin{tikzpicture}[scale=1.5]
\begin{scope}[shift={(-1, 0)}]
\node[state, user2] (x1) [label={[user2color]above:$\kernelfeat(\state_1)$}]    at (140:0.9) {};
\node[state, user2] (x2) [label={[user2color]below:$\kernelfeat(\state_2)$}]    at (220:0.9) {};
\end{scope}

\node[state, user1] (x)  [label={[user1color]left:$\kernelfeat(\state)$}]       at (-1, 0) {};

\node[state, user2] (y1) [label={[user2color]above:$\kernelfeat(\outstate_1)$}] at ( 60:1) {};

\node[state, user2] (y2) [label={[user2color]below:$\kernelfeat(\outstate_2)$}] at (300:1) {};

\path[edge, user2]%
	(x1) edge (y1)
	(x2) edge (y2);

\node[state, hint1] (h) at (0.6532, 0) {};

\path[edge, hint1] (x) edge node[above, hint1color] {$\pol_\text{GPR}(\state)$} (h);
\end{tikzpicture}
\end{center}
\caption{The embedding of the trace states (blue) and the student state (red) from the left into
the edit distance space via the embedding
$\kernelfeat$. The recommendation of the Gaussian Process Regression (GPR) policy 
$\pol_\text{GPR}(\state)$
for the current student state $\state$ is shown in orange.}
\label{fig:chf_edit_distance_space}
\end{subfigure}
\hspace{0.03\textwidth}
\begin{subfigure}[t]{0.3\textwidth}
\begin{center}
\begin{tikzpicture}[scale=1.5]
\begin{scope}[shift={(-1, 0)}]
\node[user2color]   (x1) [label={[user2color]above:$\state_1$}]    at (140:0.9) {a};
\node[user2color]   (x2) [label={[user2color]below:$\state_2$}]    at (220:0.9) {b};
\end{scope}

\node[user1color]   (x)  [label={[user1color]left:$\state$}]       at (-1, 0) {ab};

\node[defaultcolor] (aa)                                           at (120:1) {aa};
\node[user2color]   (y1) [label={[user2color]above:$\outstate_1$}] at ( 60:1) {aac};

\node[defaultcolor] (bb)                                           at (240:1) {bb};
\node[user2color]   (y2) [label={[user2color]below:$\outstate_2$}] at (300:1) {bbc};

\node[defaultcolor] (h) at (1, 0) {abc};

\path[edge, user2]%
(x1) edge [bend left]  (y1)
(x2) edge [bend right] (y2);

\path[edge, <->]%
(x)  edge (x1) edge (x2) edge (aa) edge (bb) edge (h)
(x1) edge (aa) edge (x2)
(aa) edge (y1)
(y1) edge (h)
(x2) edge (bb)
(bb) edge (y2)
(y2) edge (h);

\path[edge, hint1] (x) edge [bend left] node[above, hint1color] {$\edit$} (h);
\end{tikzpicture}
\end{center}
\caption{The legal move graph from the left figure, including the edit $\edit$ (orange)
which corresponds to the recommended edit of GPR from the center figure.}
\label{fig:chf_pre_image}
\end{subfigure}
\caption{An illustration of the Continuous Hint Factory on a simple dataset of strings.
First, we compute pairwise edit distances
between the student's current state (red) and trace data (blue). These edit distances correspond
to the shortest paths in the legal move graph (left). The edit distances correspond to a continuous
embedding, which we call the edit distance space (center). In this space, we can infer an optimal
edit (orange) using machine learning techniques, such as Gaussian Process regression (GPR).
Finally, we infer the corresponding hint in the original legal move graph (right), which can then
be displayed to the student.}
\label{fig:chf}
\end{figure}
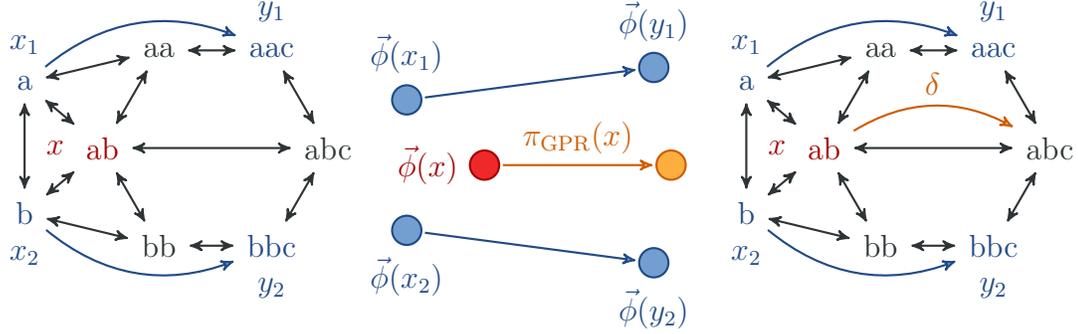

The notion of a shortest sequence of edits transforming a student's state $\lstate$ to a
solution $\rstate$ now has a proper graph-theoretical interpretation. It is the shortest
path from $\lstate$ to $\rstate$ in the legal move graph. The length of this shortest path is
a \emph{distance} $\dist$ between $\lstate$ and $\rstate$. As an example, consider 
once more the legal move graph in Figure~\ref{fig:chf_legal_move_graph} (right). Here we find, for 
example, $\dist(\text{ab}, \text{ab}) = 0$, $\dist(\text{ab}, \text{abc}) = 1$ and 
$\dist(\text{ab}, \text{bbc}) = 2$.

We implicitly assumed that all edges in a legal move graph have the length $1$. We can generalize 
this notion by specifying the length of each edge via an \emph{edit cost function}, which leads us 
to the concept of an \emph{edit distance}.

\begin{dfn}[Edit Cost Function and Edit Distances]
Let $\states$ be a set and $\edits$ be an edit set on $\states$.
A function $\editCost : \edits \times \states \to \R^+$ is called an \emph{edit cost function} on
$\edits$. We call $\editCost(\edit, \state)$ the \emph{cost} of applying edit $\edit$ to the state
$\state$.

We call an edit cost function \emph{symmetric} if $\editCost(\edit, \state) = \editCost(\edit^{-1}, \edit(\state))$
for all states $\state \in \states$, all edits $\edit \in \edits$, and all inverse edits
$\edit^{-1}$ for $\edit$ on $\state$.

Let $\graph_{\states, \edits}$ be the legal move graph according to $\states$ and $\edits$ and let
$\editCost$ be an edit cost function on $\edits$. The \emph{edit distance} $\dist_{\edits, \editCost}$
according to $\edits$ and $\editCost$ is defined as the shortest path distance in the legal move
graph $\graph_{\states, \edits}$ with the edge weights $\elabel(\lstate, \rstate) =
\min_{\edit \in \edits} \{ \editCost(\edit, \lstate) | \edit(\lstate) = \rstate \}$.
So the formula for the edit distance is:
\begin{equation}
\dist_{\edits, \editCost}(\lstate, \rstate) := \min_{\substack{
	\edit_1, \ldots, \edit_\timelim \in \edits\\
	\state_1, \ldots, \state_\timelim \in \states
	}}
\left\{ \sum_{\timeidx=1}^\timelim \editCost(\edit_\timeidx, \state_\timeidx) \middle|
	 \state_1 = \lstate, \edit_\timeidx(\state_\timeidx) = \state_{\timeidx+1}, \edit_\timelim(\state_\timelim) = \rstate \right\}
	\label{eq:edit_dist}
\end{equation}
If no path between $\lstate$ and $\rstate$ exists, we define $\dist_{\edits, \editCost}(\lstate, \rstate) = \infty$.
\end{dfn}

Edit distances enable us to specify hint policies formally. For example, the policy of Zimmerman 
and Rupakheti selects for the input state $\lstate$ the closest correct solution $\rstate$ 
according to a given edit distance and returns the first edit on the shortest path between 
$\lstate$ and $\rstate$.

Unfortunately, not all edit distances are applicable in practice. Consider the Snap example from 
Figure~\ref{fig:snap}. In this domain, the order of many blocks in the program is insignificant
to the function of the program. Therefore, one may wish to apply an edit distance which
works on unordered trees. However, edit distances on such unordered trees are NP-hard \cite{Zhang1992},
making them infeasible in practice. The subset of efficiently computable edit distances
includes the following.
First, the string edit distance \cite{Levenshtein1965} from our example in 
Figure~\ref{fig:chf_legal_move_graph}. For this example, we can define the cost function as 
$\editCost(\del_\nodeidx, \state) = \editCost(\ins_{\nodeidx, \lnode}, \state) = 
\editCost(\rep_{\nodeidx, \lnode}, \state) = 1$ for all $\nodeidx \in \N, 
\state \in \states, \lnode \in \{\text{a}, \text{b}, \text{c}\}$. Note that $\editCost$ is
symmetric. In this case, the overall edit distance can be computed in $\effic(\nodelim^2)$ using a
dynamic programming algorithm \cite{Levenshtein1965}. This dynamic 
programming scheme can be extended to a broad class of edit distances on strings, including skip 
operations and arbitrary (metric) cost functions \cite{Giegerich2004,Paassen2016Neurocomputing}.
Such string edit distances have also been successfully applied to computer programs by representing
them as sequences of code statements or as execution traces \cite{Paassen2016Neurocomputing,Paassen2016EDM}.

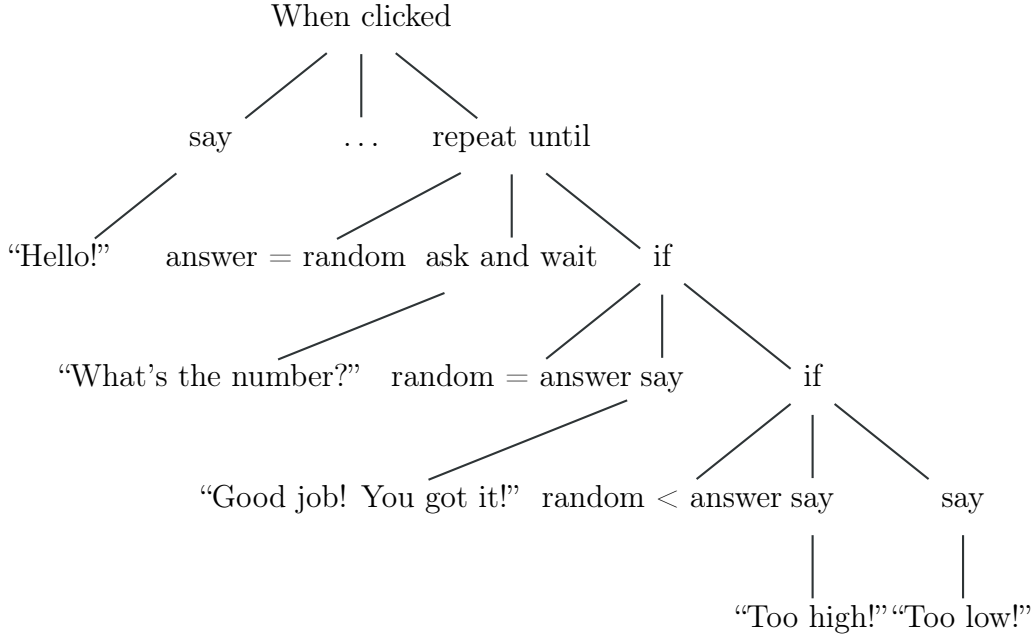
\begin{figure}
\begin{center}
\begin{tikzpicture}[every node/.append style={text height=0.2cm, text depth=0.2cm}, yscale=0.8]

\node (root)    at ( 0,  0) {When clicked};
\node (say)     at (-2, -2) {say};
\node (saystr)  at (-4, -4) {\enquote{Hello!}};
\node (dots)    at ( 0, -2) {$\ldots$};
\node (repeat)  at ( 2, -2) {repeat until};
\node (repcond) at (-1, -4) {answer = random};
\node (ask)     at ( 2, -4) {ask and wait};
\node (askstr)  at (-2, -6) {\enquote{What's the number?}};
\node (if1)     at ( 4, -4) {if};
\node (if1cond) at ( 2, -6) {random = answer};
\node (fin)     at ( 4, -6) {say};
\node (finstr)  at ( 0, -8) {\enquote{Good job! You got it!}};
\node (if2)     at ( 6, -6) {if};
\node (if2cond) at ( 4, -8) {random < answer};
\node (hi)      at ( 6, -8) {say};
\node (histr)   at ( 6, -10){\enquote{Too high!}};
\node (lo)      at ( 8, -8) {say};
\node (lostr)   at ( 8, -10){\enquote{Too low!}};

\path[edge, -]%
(root) edge (say) edge (dots) edge (repeat)
(say) edge (saystr)
(repeat) edge (repcond) edge (ask) edge (if1)
(ask) edge (askstr)
(if1) edge (fin) edge (if2)
(if1) edge (if1cond)
(fin) edge (finstr)
(if2) edge (hi) edge (lo)
(if2) edge (if2cond)
(hi) edge (histr)
(lo) edge (lostr);

\end{tikzpicture}
\end{center}
\caption{An abstract syntax tree, simplified for clarity, corresponding to the Snap program shown in 
Figure~\ref{fig:snap}.}
\label{fig:ast}
\end{figure}

Second, we note the tree edit distance of \citeN{Zhang1989} which permits deletions,
insertions, and replacements of single nodes in trees. The tree edit distance has been particularly
popular in learning environments for computer programming tasks as computer programs are oftentimes
represented by \emph{abstract syntax trees} \cite{Choudhury2016,Freeman2016,Nguyen2014,Rivers2015}.
For example, the program shown in Figure~\ref{fig:snap} would correspond to the abstract syntax tree
shown in Figure~\ref{fig:ast}. The added complexity of tree data structures compared to strings is
reflected in the higher computational complexity of the tree edit distance of $\effic(\nodelim^4)$ 
\cite{Zhang1989}.
More flexibility is offered by a two-stage approach where some special subtrees, such as functions
in a program, may be arbitrarily re-ordered and a string or tree edit distance is used to compare
the single functions \cite{Mokbel2013EDM,Price2017EDM}.
A computationally cheaper alternative has been suggested by \citeN{Zimmerman2015} who compute an edit
distance on $pq$-grams in trees, meaning small subtrees, which results in a considerably faster
runtime of $\effic(\nodelim \cdot \log(\nodelim))$ \cite{Augsten2008}.

Beyond computational complexity, a key challenge to edit distances is that they do not 
necessarily correspond to the \emph{semantic} distance between states. Consider again the Snap 
example in Figure~\ref{fig:snap}. Here, we could replace any of the strings in \enquote{say} or 
\enquote{ask} blocks with a slightly different version without changing the basic computed function 
of the program. We could also exchange the \enquote{too high} and \enquote{too low} statements in 
the program if we also replace the \enquote{random < answer} comparison with a \enquote{random > 
answer} comparison. In theory, we can apply arbitrarily many edits to a given program without 
changing the computed function. Conversely, we could also remove the \enquote{repeat until} block 
in the program, changing only a single block, and severely change the behavior of the program. This 
mismatch between edit distance and semantic distance can negatively impact the utility of generated
hints. In particular, edits may be recommended which get the student syntactically closer to a
correct solution but may be semantically irrelevant or even confusing.

One approach to address this issue is \emph{canonicalization}, which essentially transforms the raw
states in a state space $\states$ to a canonic form, such that semantically equivalent states
have the same canonic form. The edit distance is then defined between canonic forms instead of raw
states,  leading to a smaller legal move graph and edits which put stronger emphasis on semantically 
relevant changes.
Canonicalization is particularly common for computer programs, where the order of binary
relations (such as $<$) or variable names can be normalized or unreachable code can be removed
\cite{Rivers2012}. Note that canonicalization is a quite general and powerful concept. In 
particular, canonicalization enables us to apply an edit distance to any 
kind of data, as long as we can provide a canonicalization $f$ which converts this data to  
editable canonic forms. For example, \citeN{Paassen2016EDM} canonicalize computer programs by
representing them in terms of their execution trace, to which simple string edit distance
measures can be applied. As such, a canonicalization $f$ can also be seen
as an extension of an edit distance $\dist$ to a new domain by defining the edit distance $\tilde 
\dist$ on the new domain as $\tilde \dist(x, y) = \dist(f(x), f(y))$.

A challenge in canonicalization lies in the fact that edits on the canonic form may not be directly 
applicable or interpretable for students. For example, students cannot easily adapt their program 
to directly influence the program's execution in the way indicated by an edit on the execution 
trace. To address this problem, \citeN{Rivers2015} suggest aligning the edits on the canonic form with 
the student's original state in a process called \emph{state reification}.
Another challenge lies in the fact that too drastic canonicalization may remove features of the
original state for which feedback would be desirable. For example,
tutoring systems for computer programming often not only intend to teach functionally correct 
programming but also programming style, such that important stylistic differences, even though 
semantically irrelevant, need to be preserved in the canonic form 
\cite{Piech2015ICML,Choudhury2016}.

Another approach to adapting an edit distance is offered by \emph{metric learning}. Instead of mapping 
two semantically equivalent states $\lstate$ and $\rstate$ to the same canonic form, metric learning
lowers the cost of edits which transform $\lstate$ to $\rstate$, such that the edit distance 
between $\lstate$ and $\rstate$ is reduced \cite{Paassen2016Neurocomputing}. This makes it possible 
to smoothly regulate the emphasis on semantics, style, and syntax and keeps the legal move 
graph intact, such that state reification is unnecessary. On the other hand, metric learning is
limited by the neighborhood structure in the legal move graph and thus can 
not easily map equivalent states which are far apart in the legal move graph to the same point.
One potential approach is to first apply (mild) canonicalization in order to 
map distant but semantically equivalent states to the same or similar canonic forms and 
subsequently apply metric learning such that the resulting distance measure mostly takes
semantics into account but does not disregard stylistic differences entirely.

In summary, the concepts of edit set, legal move graph, and edit distance permit us to define a hint
policy that recommends the first edit on the shortest path to a solution \cite{Zimmerman2015}.
In order to ensure that such hints are helpful, we require an edit distance which roughly corresponds
to the semantic distance, but also takes important stylistic features into account.
Still, our definitions are insufficient to select helpful edits as hints reliably.
In particular, there may be many equally short paths toward a correct solution, and we do not know
which one to take. Consider the Snap programming example from Figure~\ref{fig:snap} again and assume
that a student requests help without having written any code. In that case, to get to
the closest correct solution in Figure~\ref{fig:snap}, we need to insert all code statements of the
abstract syntax tree in Figure~\ref{fig:ast}, and a priori each of these insertions is an equally
valid hint. Past research suggests that we should focus on insertions which other students have
performed in the same situation, which brings us to the notion of traces and interaction networks.

\subsection{Traces and Interaction Networks}

If we wish to capture what students have done in the past, we require a formal notion of student 
movement through the state space. This is offered by the notion of a trace suggested by \citeN{Eagle2012}%
\footnote{Note that this definition is not exactly equivalent to the one given 
by \citeN{Eagle2012}. In particular, they do not require actions to be \emph{deterministic}, that is,
in their framework the same action applied to the same state may lead to different
subsequent states. For the sake of brevity, we refrain from this probabilistic extension here.}.

\begin{dfn}[Trace]
Let $\states$ be a set and $\edits$ be an edit
set on $\states$. Then, a sequence $\state_1, \edit_1, \ldots, \edit_{\timelim-1}, \state_\timelim$
is called a \emph{trace} if for all $\timeidx$ $\state_\timeidx \in \states$, $\edit_\timeidx \in \edits$ and
$\state_{\timeidx + 1} = \edit_\timeidx(\state_\timeidx)$.
\end{dfn}

We are particularly interested in how close a trace is to a new student's solution. To answer this 
question, we need to couple the notion of a trace with the notion of the legal move graph. This 
coupling is provided by the notion of an interaction network \cite{Eagle2012} and a solution space
\cite{Rivers2014,Rivers2015}.

\begin{dfn}[Interaction Network]
Let $\states$ be a set and $\edits$ be an edit set on $\states$. Further, let
$\{ (\state^\traceidx_1, \edit^\traceidx_1, \ldots, \edit^\traceidx_{\timelim_\traceidx-1},
\state^\traceidx_{\timelim_\traceidx}) \}_{\traceidx = 1, \ldots, \tracelim}$
be a set of traces. The \emph{interaction network} corresponding to this set of traces
is defined as the graph $\graph = (\nodes, \edges)$ where
\begin{align}
\nodes &= \Big\{ \state^\traceidx_\timeidx \Big| \traceidx \in \{1, \ldots, \tracelim \},
\timeidx \in \{1, \ldots, \timelim_\traceidx\} \Big\} \\
\edges &= \Big\{ (\state^\traceidx_\timeidx, \state^\traceidx_{\timeidx+1}) \Big|
\traceidx \in \{1, \ldots, \tracelim \}, \timeidx \in \{1, \ldots, \timelim_\traceidx-1\} \Big\}
\end{align}
We also call $\nodes$ a \emph{solution space}.
\end{dfn}

Consider the example shown in Figure~\ref{fig:chf_legal_move_graph}, which shows two
traces in blue. These traces cover the strings \enquote{a,} \enquote{aac,} 
\enquote{b,} and \enquote{bbc}. Therefore, the interaction network for this case would
only contain these four strings and the edges (\enquote{a,} \enquote{aac}) as well as
(\enquote{b,} \enquote{bbc}).

Ideally, the edit set used by students exactly corresponds to the edit set of the legal move graph, but
there are many cases where this condition may not hold. For example, the edit sets might
be different because students work on a different representation compared to the representation used
to compute edit distances, for example, due to canonicalization \cite{Rivers2012,Rivers2015}.
Another reason may be that student's states may be recorded only at certain points in
time (e.g., when they explicitly hit a \enquote{save} button) such that multiple actions may have
been performed since the last recorded state. Finally, the concrete actions of a student
may not be available because the students work on the task off-line and submit their
current state only if they need help from the system.
In these cases, we have to assume that students may \enquote{jump} in the legal move graph from
state to state and the exact path they have taken needs to be inferred by the system 
\cite{Piech2015}.

With a formal notion of the actions of past students, we have now aggregated all concepts we need to
provide an integrated view of existing hint policies in the literature.

\subsection{Hint policies}

Recall the definition of a hint policy (Definition~\ref{dfn:hint_policy}). We are referring to a
function which outputs an edit for each possible input state. In the remainder of this
section, we are going to provide a short review of the hint policies that have been suggested
in the literature.

The arguably simplest policy is the one of \citeN{Zimmerman2015}, which always
recommends the first edit on the shortest path to the closest solution.
Such an approach does not even require student data, except for at least one example of a
correct solution of the task. A drawback of the Zimmerman policy is that it does not consider
whether the edits towards the closest correct solution correspond to critical steps toward a
solution or relatively unimportant stylistic differences. \citeN{Rivers2015} address this
issue in their Intelligent Teaching Assistant for Programming (ITAP), an intelligent tutoring
system for Python programming. Their technique involves the following steps:
First, they retrieve the closest solution according to the tree edit distance on canonic forms. 
Second, they use the edits which transform the student state into the closest correct solution to 
construct intermediate states. Third, of these intermediate states, the one with the 
highest desirability score is selected for feedback, where the desirability score is a weighted sum 
of the frequency in past student data, the distance to the student's state, the number of 
successful test cases the state passes, and the distance to the solution \cite{Rivers2015}. Finally, 
an inverse canonicalization (state reification) step is applied to infer edits that can be directly
applied to the student's state to transform it to the selected state. This approach has been shown 
to provide helpful edits in almost all cases for a broad range of tasks \cite{Rivers2015}.
Note that the success of the Rivers policy hinges upon meaningful frequency information. If no or
little frequency information is available, the hints provided by the Rivers policy may not be
representative of generic steps toward a solution but rather of specificities of the reference
solution that was selected.

The approaches of \citeN{Zimmerman2015} as well as \citeN{Rivers2015} have been part of a 
study by \citeN{Piech2015} who compared several hint policies
with expert recommendations on a large-scale dataset consisting of over a million states
from the \emph{Hour of Code} Massive Open Online Course (MOOC). The data was collected from two
beginner's programming tasks in a block-based programming language.
They found that the policy which agreed most with the recommendations of tutors was a variant of 
the Zimmerman policy, in which the cost of an edit $\edit$ was determined based on the inverse frequency 
of the target state in the dataset, that is, edits which lead to less frequent states were considered
more expensive \cite{Piech2015}. This approach makes states appear closer when they can be reached by
crossing states that have been visited often. Note that this approach critically relies on
frequency information, which may not be available in sparsely populated spaces,
where almost no state is visited more than once.

An alternative approach to approaching the closest correct solution directly is to guide students
along a trace, as proposed by \citeN{Gross2015AIED} in the \emph{JavaFIT} system\footnote{\url{https://javafit.de/}}.
They distinguish between two types of help-seeking behavior, namely searching for errors or 
searching for a next-step. In both cases, they retrieve the closest state to the student's state in
the interaction network. If students are trying to find an error in their code,
the system recommends an edit leading the student toward this reference state,
thereby attempting to correct the error.
If students assume that their current state is correct, but they are looking for a next step,
the system recommends an edit toward the \emph{successor} of the reference state,
thereby guiding the student closer to a solution \cite{Gross2015AIED}.
This policy can be seen as an instance of \emph{case-based reasoning}, where recommendations are
based on a similar case from an underlying case base. \citeN{Freeman2016} have taken this view to
analyze Python programs and used a weighted tree edit distance to retrieve similar cases.
Also similar to case-based reasoning, \citeN{Gross2014} proposed example-based feedback, in which
the closest prototypical state in a dataset is retrieved and shown to the student to
elicit self-reflection and sense-making in order to improve their own state.
If the closest state in the case base is sufficiently similar to the student's state
and corresponds to a capable student, such an approach can provide hints which emulate the actions
of a capable \enquote{virtual twin} of the student. However, if only few reference solutions exist,
the selected next state may still be fairly dissimilar and edits toward the next state may include
not only error-correcting hints or next-step hints but also stylistic or strategic choices which do
not correspond to the student's goals.

\citeN{Lazar2014} propose yet a different approach by applying edits that have been
frequent in past student traces to manipulate the current student's state until an
edit is found that achieves better unit test scores. As with the Piech policy,
the policy of \citeN{Lazar2014} critically relies on frequency information, albeit for edits
instead of states, which may not always be available. Furthermore, edits which may be generally
important for a task may not necessarily be helpful in a specific situation.

An alternative view is provided by the Hint Factory, which analyzes the question of choosing the
optimal edit in a given state in a mathematically precise fashion via Markov Decision Processes
\cite{Barnes2008}. In particular, the Hint Factory always returns the edit which maximizes the
expected future reward, where a reward is given whenever a student has achieved a correct solution.
The Hint Factory was originally created as a
hint-generation add-on to the \emph{DeepThought} instruction system for deductive logic
\cite{Barnes2008}. Several studies have demonstrated that the Hint Factory reduces student dropout
and helps students to complete more problems more efficiently \cite{Stamper2012,Eagle2013}.
The Hint Factory has also been applied to further domains, such as the serious game BOTS
\cite{Hicks2014} or the SNAP programming environment \cite{Price2017}.

Note that the Markov Decision Process model relies on an estimate of the transition probability
distribution $\prob(\state' | \state, \edit)$ of moving to state $\state'$ from $\state$ via the
edit $\edit$. The Hint Factory estimates this probability distribution based on transition
frequencies in the trace data and therefore requires meaningful frequency information.
As such, the Hint Factory can provide hints only for states that are part of the interaction network,
and for which a directed path to a correct solution in the interaction network exists.
This has been dubbed the \emph{hintable subgraph} \cite{Barnes2016}.
In practice, students may move outside the hintable subgraph. Indeed, research has shown that for a
reasonably small, open-ended programming task, over 90\% of states are visited only once, indicating
that future students will likely visit states that have not been seen before and may not even be
connected to previously seen states in the legal move graph \cite{Price2015}. Also note that the
number of unique states remained high even after applying harsh canonicalization \cite{Price2015}.
This result matches our own two datasets, where 97.23\% and 82.79\% of states were visited only
once. So how can the Hint Factory be extended to such sparsely populated state spaces?
A first approach has been proposed by \citeN{Price2016}, who suggest
\emph{contextual tree decomposition} (CTD) which generates interaction networks only for small
subtrees of the students' abstract syntax trees. Due to the size limitation,
the state space for each subtree is significantly smaller and thus more densely populated with
student data. However, the approach faces an ambiguity challenge in that one 
hint is generated for each (small) subtree of the student's state, and the student or the system 
has to select from these possible hints \cite{Price2017EDM}.

Overall, we observe that all previous approaches are either limited by their reliance on frequency 
data, namely the Hint Factory, the Piech policy, and the Lazar policy, or by generating hints
based only on a single reference solution, namely the Zimmerman policy, the Gross policy, or
the Rivers policy. Our approach is an attempt to generate hints based on \emph{multiple} reference
solutions, but without relying on frequency information. More specifically, we use a weighted
average of multiple reference solution to express a virtual state to which the student should move,
and we select the weights for this average such that the resulting virtual state corresponds to
the probabilistically optimal next state of a capable student. As such, we use the same basic
approach as the Hint Factory, in that we also try to bring the student closer to the next state
of capable students in the same situation. However, we extend the Hint Factory by basing our
prediction not on frequency counting, but on the movements of students in \emph{similar} situations
through the space of possible solutions. This state of possible virtual solutions, expressible
as weighted averages of states we have seen before, is continuous; hence the name
\emph{Continuous Hint Factory} (CHF).

Note that embedding states in a continuous state has already been proposed by \citeN{Piech2015ICML},
who constructed such an embedding via neural networks.
The embedding is computed by executing the programs on example data and recording
the variable states $P$ before executing a block of code $A$ as well as the variable
states $Q$ after $A$ has been executed.
Both $P$ and $Q$ are embedded in a common space via a single-layer
neural network, yielding the representations $f_P$ and $f_Q$. Then, a matrix
$M_A$ is constructed which maps $f_P$ to $f_Q$, that is, $M_A$ is constructed such that 
$f_Q \approx M_A \cdot f_P$. This matrix $M_A$ is the embedding of the code block $A$ 
\cite{Piech2015ICML}. However, this work has two crucial limitations: First, it relies on a
task-specific representation of $f_P$ and $f_Q$, which is generated via execution, whereas the CHF
only relies on edit distances, which are not task-specific \cite{Mokbel2013EDM}.
Second, we provide a technique to convert the predictive result in the continuous space to an
actual, human-readable edit, which the Piech approach lacks.

We also note connections to other approaches cited before. First, the CHF is connected to the work 
of \citeN{Gross2015AIED}, in that we also recommend following the actions of students in a similar 
situation, but we integrate knowledge of more than one student. Second, similar to the work of 
\citeN{Lazar2014}, we recommend edits which are frequent in past student data but focus on those 
edits which have been applied in similar states. Finally, we incorporate many of the key concepts 
and approaches of \citeN{Rivers2015}, in that we also apply canonicalization, and build upon 
the concept of path construction, a desirability score, as well as state reification to infer an
edit which corresponds to the optimal hint in the embedding space.
However, we extend this approach by considering not only edits toward the closest correct solution
but edits toward all reference solutions and by replacing their desirability score with 
the distance to the recommended next state in the edit distance space. This alternative score
incorporates the spirit of many of the criteria proposed by \citeN{Rivers2015}, as it also
punishes going too far away from the student's current solution, rewards getting closer to the goal,
and represents what other students generally did, but it relies neither on frequency information,
nor on an expert-chosen weighting between the different criteria.

In the next section, we introduce the CHF in more detail.

\section{Continuous Hint Factory}
\label{sec:chf}

The goal of the Continuous Hint Factory (CHF) is to identify a state which represents where 
students should move next on their way towards a correct solution and to recommend an edit which 
gets as close as possible to this selected state. In doing so, we also want to be able to select 
next states which are not contained in the data of past data but can be represented as mixtures of 
previously seen states.

To implement this goal, the CHF involves three steps. First, we embed past student data in a 
continuous space by means of an edit distance. In this space, we can represent any mixture of known
states as a vector. Second, we develop a hint policy in this embedding space based on Gaussian
process prediction for structured data \cite{Paassen2017NPL}. This policy returns what a capable
student would do in the respective input state, represented as a mixture of states from traces of
such capable students. In a final step, we transform this mixture into a human-readable edit by
selecting  the edit which brings us closer to all states with positive mixture coefficients and
further away from all states with negative mixture coefficients.

In the remainder of this section, we describe each of these three steps - embedding in the edit 
distance space, prediction, and pre-image identification - in turn.

\subsection{The Edit Distance Space}

In a first step, we embed the states of past students in a Euclidean vector space, such that the 
edit distances between any two states $\lstate$ and $\rstate$ corresponds to the Euclidean distance 
between the corresponding vectors $\kernelfeat(\lstate)$ and $\kernelfeat(\rstate)$. Such a space 
is necessary to ensure that \enquote{mixing} states becomes meaningful. If we have vectors 
$\kernelfeat(\lstate)$ and $\kernelfeat(\rstate)$ which correspond to $\lstate$ and $\rstate$, we 
can easily mix those two vectors by adding them or subtracting them from each other.

A simple example of such an embedding is shown in Figure~\ref{fig:chf_edit_distance_space} (left).
In this figure, the strings \enquote{a}, \enquote{b}, \enquote{ab}, \enquote{aa}, \enquote{aac}, 
\enquote{bb}, \enquote{bbc}, and \enquote{abc} are embedded in a two-dimensional space, namely the 
two dimensions of the paper. The Euclidean distance between the strings on the paper
approximately corresponds to their edit distance, that is, strings with an edit distance
of $2$, such as \enquote{ab} and \enquote{aac}, are about twice as far away from each other
on the paper compared to strings with an edit distance of $1$, such as \enquote{ab} and 
\enquote{aa}.
Note that this embedding is not exact. For example, the distance on the paper between
\enquote{a} and \enquote{b} is much larger compared to the distance between \enquote{a}
and \enquote{ac}, even though in both cases the edit distance is $1$.
Another example is shown in Figure~\ref{fig:snap_embedding}. This figure displays ten traces of 
students working on the guessing game task from Figure~\ref{fig:snap}, such that the 
distance between states roughly corresponds to the edit distance between them.
In both cases we observe that finding a vectorial embedding, such that the edit distance 
corresponds exactly to the Euclidean distance in the embedding, is not trivial. Importantly,
though, theoretical results show that such an embedding does always exist.

\begin{thm}[Latent Distance Space] \label{thm:latent_space}
Let $\states$ be some finite set and $\dist : \states \times \states \to \R$ be a
function such that for all $\lstate, \rstate \in \states$ it holds:
$\dist(\lstate, \lstate) = 0$, $\dist(\lstate, \rstate) \geq 0$ and $\dist(\lstate, \rstate)
= \dist(\rstate, \lstate)$. Then, there exists a vector space $\kernelspace \subset \R^\dims$
for some $\dims \in \N$, and a mapping $\kernelfeat : \states \to \kernelspace$, such that for all 
$\lstate, \rstate \in \states$:
\begin{equation}
	\dist(\lstate, \rstate)^2 = \transp{(\kernelfeat(\lstate) - \kernelfeat(\rstate))} \cdot 
\Lambda 
\cdot (\kernelfeat(\lstate) - \kernelfeat(\rstate))
\end{equation}
where $\Lambda$ is a diagonal matrix with entries in $\{-1, 0, 1\}$.
\begin{proof}
Refer to Theorem 1 in \citeN{Hammer2010} as well as page 122 in \citeN{Pekalska2005}.
\end{proof}
\end{thm}

Note that if $\Lambda$ has no negative entries, $\dist$ corresponds to the Euclidean distance
in the embedding space. Otherwise, there exist points in the latent vector space
$\kernelspace$ for which the pairwise distance becomes \emph{negative}, which may cause errors
in subsequent processing. These negative entries can be addressed by various \emph{eigenvalue 
correction} techniques, such as setting the negative entries to zero (clip eigenvalue 
correction), replacing them with their absolute value (flip eigenvalue correction) or adding an 
offset to $\Lambda$, such that all entries become positive \cite{Gisbrecht2015}. Note that this 
correction is an \emph{approximation} and does distort the distances, but only to the extent to 
which negative entries are present.

Based on this embedding, we can prove the main theorem of our work, namely that for any
symmetric edit distance we can find an Euclidean embedding, which we call
the edit distance space.

\begin{thm}[Edit Distance Space] \label{thm:edit_distance_space}
Let $\nodes \subset \states$ be a solution space taken from a state space $\states$, let $\edits$ be a 
symmetric edit set on $\states$ and let $\editCost : \edits \times \states \to \R^+$ be a
symmetric \emph{edit cost function} on $\edits$. Further, let $\distmat$
be the eigenvalue-corrected version of the distance matrix with entries $\distmat_{\ldataidx 
\rdataidx} = \dist_{\edits, \editCost}(\state_\ldataidx, \state_\rdataidx)^2$ for all 
$\state_\ldataidx, \state_\rdataidx \in \states$.

Then, there exists a vector space $\kernelspace = \R^\dims$ for some $\dims \in \N$
which we call the \emph{edit distance space} for $\dist_{\edits, \editCost}$; and there exists a 
mapping $\kernelfeat : \states \to \kernelspace$, such that for all $\lstate, \rstate \in \nodes$
it holds: $\lVert \kernelfeat(\lstate) - \kernelfeat(\rstate) \rVert_2^2 = \distmat(\lstate, 
\rstate)$,
i.e.\ the Euclidean distance in $\kernelspace$ corresponds to $\dist_{\edits, \editCost}$, up to 
eigenvalue correction.
\begin{proof}
We first show that, under our constraints on $\edits$ and $\editCost$, the resulting edit distance
$\dist_{\edits, \editCost}$ fulfills the constraints of Theorem~\ref{thm:latent_space}.
\begin{description}
\item[$\dist_{\edits, \editCost}(\lstate, \rstate) \geq 0$:] If $\lstate$ and $\rstate$ are 
connected in the legal
move graph, $\dist_{\edits, \editCost}(\lstate, \rstate)$ is a sum of non-negative contributions
(because $\editCost$ is non-negative), and thus $\dist_{\edits, \editCost}(\lstate, \rstate) \geq 
0$.
Otherwise $\dist_{\edits, \editCost}(\lstate, \rstate) = \infty > 0$.
\item[$\dist_{\edits, \editCost}(\lstate, \lstate) = 0$:] For all $\lstate$ we can use the empty
edit sequence to transform $\lstate$ to $\lstate$. The cost of the empty edit sequence is $0$,
independent of the cost function $\editCost$. As we have shown that $\dist_{\edits, 
\editCost}(\lstate, \lstate) \geq 0$, there can also be no cheaper edit sequence. Therefore,
we obtain $\dist_{\edits, \editCost}(\lstate, \lstate) = 0$
\item[$\dist_{\edits, \editCost}(\lstate, \rstate) = \dist_{\edits, \editCost}(\rstate, \lstate)$:] 
Let
$\lstate, \rstate \in \states$ such that $\lstate$ and $\rstate$ are connected in the legal move 
graph.
Let $\edit_1, \ldots, \edit_\timelim$ be a sequence of edits that transforms $\lstate$ to $\rstate$
such that the cost is minimal.
Because $\edits$ is symmetric, we can construct the sequence of edits $\edit_\timelim^{-1},
\ldots, \edit_1^{-1}$ which transforms $\rstate$ to $\lstate$. Because $\editCost$ is symmetric
we know that the cost of this path is equal to the cost of $\edit_1, \ldots, \edit_\timelim$,
which in turn implies $\dist_{\edits, \editCost}(\lstate, \rstate) \geq \dist_{\edits, 
\editCost}(\rstate, \lstate)$.
We can do the same argument in the other direction (from $\rstate$ to $\lstate$), such that
$\dist_{\edits, \editCost}(\lstate, \rstate) \leq \dist_{\edits, \editCost}(\rstate, \lstate)$,
which implies $\dist_{\edits, \editCost}(\lstate, \rstate) = \dist_{\edits, \editCost}(\rstate, 
\lstate)$.
If there is no path from $\lstate$ to $\rstate$ in the legal move graph, then there is also no
path from $\rstate$ to $\lstate$, and it holds $\dist_{\edits, \editCost}(\lstate, \rstate) = \infty 
= \dist_{\edits, \editCost}(\rstate, \lstate)$.
\end{description}
Theorem~\ref{thm:latent_space} now yields the required embedding. Because of eigenvalue correction,
this embedding is Euclidean.
\end{proof}
\end{thm}

Note that the construction of the edit distance space depends on example data. How we select this 
example data is crucial for a viable edit distance space and, in turn, for a helpful hint policy. 
We suggest selecting example data with two heuristics. First, we should limit ourselves to data of 
successful students, meaning data of students who did end up in a correct solution to the task. 
Second, we should incorporate the goal-directedness criterion suggested by 
\citeN{Rivers2014}, that is, we should only incorporate those intermediate solutions which get 
closer to the correct solution the student ended up in.

In our approach, we make extensive use of the edit distance space. In particular, we replace the 
problem of finding a hint policy for the original edit set of the edit distance by finding a hint 
policy in the edit distance space.

\subsection{A Hint Policy in the Edit Distance Space}

Due to Theorem~\ref{thm:edit_distance_space} we know that, for a symmetric edit
distance $\dist_{\edits, \editCost}$, there exists an embedding in a vector space
$\kernelspace \subset \R^\dims$, such that the edit distance corresponds to the
Euclidean distance in $\kernelspace$ after eigenvalue correction. The main advantage of the edit 
distance space $\kernelspace$ is that constructing a hint policy for vectors is much easier 
compared to a hint policy for arbitrary states. In particular, we can define edits in a vector 
space as vectors which are added to the input states. Consider the example edit distance space in 
Figure~\ref{fig:chf_edit_distance_space}. In this space, string edits become vectors, displayed 
here as blue and orange arrows, and the resulting state after applying the edit corresponds to 
adding the vector to the original state.

In formal terms, we define the edit set $\edits_\kernelspace$ and the cost function 
$\editCost_\kernelspace$ in the edit distance space as follows.
\begin{equation}
\edits_\kernelspace = \{ \edit_\editvec | \editvec \in \kernelspace \} \quad \text{where} \quad 
\forall \kernelfeat \in \kernelspace : \edit_\editvec(\kernelfeat ) = \kernelfeat  + \editvec \quad 
\text{and} \quad
\editCost_\kernelspace(\edit_\editvec, \kernelfeat ) = \lVert \editvec \rVert
\end{equation}

The edit distance resulting from this definition of edit set $\edits_\kernelspace$
and edit cost function $\editCost_\kernelspace$ in the edit distance space is provably equivalent 
to the original edit distance, up to eigenvalue correction.

\begin{thm}
Let $\nodes \subset \states$ be some solution space from a state space $\states$, let $\edits$ be a 
symmetric edit set on $\states$ and let $\editCost : \edits \times \states \to \R$ be a
symmetric \emph{edit cost function} on $\edits$. Further, let $\distmat$
be the eigenvalue-corrected version of the matrix with entries $\distmat_{\ldataidx 
\rdataidx} = \dist_{\edits, \editCost}(\state_\ldataidx, \state_\rdataidx)^2$ for all 
$\state_\ldataidx, \state_\rdataidx \in \nodes$. Finally, let 
$\dist_\kernelspace(\state_\ldataidx, 
\state_\rdataidx)$ be the edit distance in the edit distance space according to 
$\edits_\kernelspace$ and $\editCost_\kernelspace$.

Then, for all $\state_\ldataidx, \state_\rdataidx \in \nodes$ it holds:
$\distmat_{\ldataidx, \rdataidx} = \dist_\kernelspace(\kernelfeat(\state_\ldataidx), 
\kernelfeat(\state_\rdataidx))^2$, i.e.\ the edit distance in the edit distance space is 
equivalent to the original edit distance $\dist_{\edits, \editCost}$, up to eigenvalue correction.

\begin{proof}
Recall that Theorem~\ref{thm:edit_distance_space} already shows that $\distmat_{\ldataidx, 
\rdataidx}$ corresponds to the squared Euclidean distance $\lVert \kernelfeat(\lstate) - 
\kernelfeat(\rstate) \rVert_2^2$. It remains to show that the edit distance in the edit distance 
space $\kernelspace$ is equivalent to the Euclidean distance in the edit distance space as well.

Let $\kernelfeat(\lstate), \kernelfeat(\rstate) \in \kernelspace$. Then, the edit 
$\edit_{\kernelfeat(\rstate) - \kernelfeat(\lstate)}$ is in $\edits_\kernelspace$.
Therefore, $\dist_\kernelspace(\kernelfeat(\lstate), \kernelfeat(\rstate))$
is at most $\lVert \kernelfeat(\rstate) - \kernelfeat(\lstate) \rVert$. Due to the triangular
inequality on the Euclidean distance we also know that there exists no point $\kernelfeat(z) 
\in \kernelspace$, such that $\lVert \kernelfeat(\lstate) - \kernelfeat(z) \rVert +
\lVert \kernelfeat(z) - \kernelfeat(\rstate) \rVert < \lVert \kernelfeat(\lstate) - 
\kernelfeat(\rstate) \rVert$. Therefore, there can exist no sequence of edits which is cheaper 
than $\lVert \kernelfeat(\rstate) - \kernelfeat(\lstate) \rVert$, which concludes the 
proof.
\end{proof}
\end{thm}

\begin{figure}
\begin{subfigure}[b]{0.5\textwidth}
\begin{center}
\begin{tikzpicture}
\begin{axis}[
	legend pos=north east,
	legend cell align=left,
	width=\textwidth,
	xmin=0.1, xmax=0.5,
	ymin=-0.15, ymax=0.25
]
\foreach \t in {1,2,...,10} {
\addplot[
quiver={
	u=\thisrow{deltax},
	v=\thisrow{deltay}
},
user2color, -stealth, semithick, forget plot]
table[x=x, y=y]
{results_data/snap_trajectory\t.csv};
}
\addplot[user1, only marks, mark=*] table[x=x, y=y] {results_data/snap_trajectory_hints.csv};
\addlegendentry{user}
\addplot[quiver={u=\thisrow{1nn_x},v=\thisrow{1nn_y}},
hint2color,-stealth,semithick]
table[x=x, y=y] {results_data/snap_trajectory_hints.csv};
\addlegendentry{1-NN}
\addplot[quiver={u=\thisrow{kr_x},v=\thisrow{kr_y}},
hint3color,-stealth,semithick]
table[x=x, y=y] {results_data/snap_trajectory_hints.csv};
\addlegendentry{NWR}
\addplot[quiver={u=\thisrow{gpr_x},v=\thisrow{gpr_y}},
hint1color,-stealth,semithick]
table[x=x, y=y] {results_data/snap_trajectory_hints.csv};
\addlegendentry{GPR}
\end{axis}
\end{tikzpicture}
\end{center}
\caption{An excerpt of the embedding for the Snap guessing game task from
Figure~\ref{fig:snap_embedding} (blue), a hypothetical new student state (red) and the
outputs of a one-nearest neighbor hint policy (1-NN), a Nadaraya-Watson kernel regression
hint policy (NWR) and a Gaussian Process hint policy (GPR) for this state, based on
a length scale of $\bandwidth = 0.5$.}
\label{fig:snap_hints}
\end{subfigure}
\hspace{0.05\textwidth}
\begin{subfigure}[b]{0.4\textwidth}
\begin{tikzpicture}
\begin{axis}[
disabledatascaling,
xlabel={$\dist(\lstate, \rstate)$},
ylabel={$\kernel_{\bandwidth, \dist}(\lstate, \rstate)$},
extra x ticks={1},
extra x tick style={grid=major, major grid style={color=scarletred3}},
extra x tick labels={},
width=\textwidth,
xmin=0,ymin=0
]
\addplot[skyblue3,thick,domain=0:3,samples=100]%
{exp(-0.5*x*x)};
\node[scarletred3, right] at (1, 0.75) {$\bandwidth$};
\end{axis}
\end{tikzpicture}
\caption{The output of the radial basis function (RBF) kernel (y-axis) for distances between two
states $\lstate$ and $\rstate$ in the range $[0, 3]$
(x-axis) and a length-scale of $\bandwidth = 1$ (red line).
\vspace{1cm}}
\label{fig:rbf}
\end{subfigure}
\caption{Two figures illustrating regression techniques (left) and the radial basis function
kernel (right).}
\end{figure}
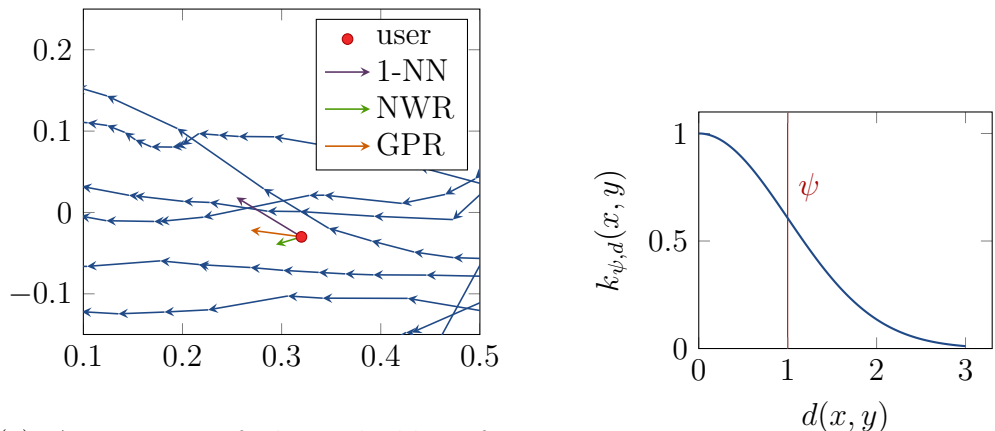

Thanks to this theorem we can replace a hint policy in the original space with a hint policy in the 
edit distance space. We can infer such a hint policy based on example data of successful students. 
In particular, assuming a dataset of traces of successful students, we denote any state in one of 
these traces as $\state_\dataidx$ and the next state in the trace as $\outstate_\dataidx$. If 
$\state_\dataidx$ is the final solution of a trace, then $\outstate_\dataidx = \state_\dataidx$.
Constructing a hint policy in the edit distance space corresponds to finding a function which 
outputs for any $\kernelfeat(\state_\dataidx)$ the edit vector
$\kernelfeat(\outstate_\dataidx) - \kernelfeat(\state_\dataidx)$. In machine learning terms, 
this is a \emph{regression problem}. The simplest approach to this problem would be one-nearest 
neighbor regression (1-NN), which looks for the closest data point in the database and returns the 
edit that has been done for that point, that is, $\pol_\text{1NN}(\kernelfeat(\state)) = 
\editvec_\dataidx$ where $\dist(\state, \state_\dataidx)$ is minimal. Unfortunately, 
such a policy tends to overestimate the importance of particularities of individual traces. 
Consider the example shown in Figure~\ref{fig:chf_edit_distance_space}.
We wish to provide a hint for point $\kernelfeat(\state)$, shown in red. Our trace data
consists of the points $\state_1$ and $\outstate_1$ as well as $\state_2$ and $\outstate_2$.
Both $\state_1$ and $\state_2$ are equally close to $\state$. Still, one-nearest neighbor has to
arbitrarily choose between both points and predict the edit corresponding to only one. 
In practical examples, we observe similar problems. Consider the excerpt of 
the Snap dataset shown in Figure~\ref{fig:snap_hints} and assume we want to provide feedback for 
a state located at the red point. In that case, a one-nearest neighbor policy would 
recommend a movement in upwards direction which is not representative of the overall movement in 
the data, which rather goes straight to the left.

We can address the overemphasis of individual particularities by integrating information from 
multiple student traces. In particular, we propose to construct a hint policy which returns a 
\emph{weighted average} of all past edits, that is
$\pol(\state) = \sum_{\dataidx = 1}^\datalim \gpcoeff_\dataidx(\state) \cdot \editvec_\dataidx$,
where $\gpcoeff_\dataidx(\state) \in \R$ are numeric weights. These weights should result in a 
compromise between past data, such that individual particularities are averaged out, and general 
trends in the data are emphasized. In the example in Figure~\ref{fig:chf_edit_distance_space}, we 
could simply set the weights for both traces to an equal value, resulting in a viable edit
(shown in orange).

The key question in such an averaging approach is how to obtain the weights
$\gpcoeff_\dataidx(\state)$. The simplest way is to scale $\gpcoeff_\dataidx(\state)$ inversely to 
the distance $\dist(\state, \state_\dataidx)$ between $\state$ and $\state_\dataidx$, i.e., such
that states $\state_\dataidx$ which are close to $\state$ are emphasized.
For that purpose, we can define $\gpcoeff_\dataidx(\state)$ as $\kernel(\state, \state_\dataidx)$
divided by $\sum_{\rdataidx = 1}^\datalim \kernel(\state, \state_\rdataidx)$ where $\kernel$ may be any
function which decreases monotonically with rising edit distance $\dist(\state, \state_\dataidx)$.
A particularly popular choice is the radial basis function (RBF) or Gaussian function which is 
defined as $\kernel_{\bandwidth, \dist}(\lstate, \rstate) = \exp(-0.5 \cdot 
\frac{\dist(\lstate, \rstate)^2}{\bandwidth^2})$ \cite{Rasmussen2005}. The RBF assigns a value of 
$1$ for a distance of $0$ and assigns lower values for higher distance, quickly approaching $0$. 
$\bandwidth$ is a hyper-parameter called \emph{length-scale} which controls the distance value at 
which $\kernel_{\bandwidth, \dist}$ reaches its maximum slope (see Figure~\ref{fig:rbf}).

This way of setting $\gpcoeff_\dataidx(\state)$ is called \emph{Nadaraya-Watson
kernel regression} (NWR) and oftentimes yields more accurate predictions compared to simple 
1-NN \cite{Paassen2017NPL}. However, NWR is highly sensitive to the choice 
of $\bandwidth$. In particular, for small $\bandwidth$ values, NWR is equivalent to 1-NN, and 
for large $\bandwidth$ values, NWR overemphasizes distant states, as is visible in 
Figure~\ref{fig:snap_hints}. Here, NWR recommends a downward 
motion, which does not reflect the movement in the local environment. An alternative to NWR is
\emph{Gaussian Process Regression} (GPR), which is well-justified via probability theory
\cite{Rasmussen2005}, and has been shown to be more accurate in prediction \cite{Paassen2017NPL}.
To apply GPR, the function $\kernel$ needs to be a \emph{kernel function}. For an exact definition 
of kernel functions, we point to the work of \citeN{Rasmussen2005}. For our 
purposes here, it suffices to state that the RBF is one such kernel function.

Let now $\kernelvec(\state)$ be the row vector of kernel values $\kernel(\state, \state_\dataidx)$ 
for all $\dataidx = 1, \ldots, \datalim$ and let $\kernelmat$ be the matrix of pairwise
kernel values $\kernel(\state_\ldataidx, \state_{\rdataidx})$ for all $\ldataidx = 1, \ldots, 
\datalim$ and $\rdataidx = 1, \ldots, \datalim$.
Then, for the point $\kernelfeat(\state)$, we define the result of the GPR hint policy as
$\pol_\text{GPR}(\state) = \sum_{\dataidx = 1}^\datalim \gpcoeff_\dataidx(\state) \cdot \editvec_\dataidx$,
where the vector $\vec \gpcoeff(\state) = (\gpcoeff_1(\state), \ldots, \gpcoeff_\datalim(\state))$ 
is computed as follows \cite{Paassen2017NPL}.
\begin{equation}
\vec \gpcoeff(\state) = \kernelvec(\state) \cdot \Big(\kernelmat + \nDevNoise^2 \cdot 
\eye^\datalim\Big)^{-1} \label{eq:gpr_mean}
\end{equation}
and where $\nDevNoise$ is a hyper-parameter which quantifies the assumed amount of noise in
our data and $\eye^\datalim$ is the $\datalim \times \datalim$ identity matrix.
Increasing $\nDevNoise$ tends to decrease the accuracy of GPR but enhances smoothness 
in the predictions as well as numerical stability \cite{Rasmussen2005}.

We highlight two key properties of GPR.
First, if $\state$ is equal to some state $\state_\dataidx$ in the recorded trace data, then
$\kernelvec(\state)$ equals the $\dataidx$th column in the matrix $\kernelmat$. The product
$\kernelvec(\state) \cdot \kernelmat^{-1}$ is then equal to a vector of zeros which is only one
at position $\dataidx$. So for $\nDevNoise = 0$, the GPR policy will return exactly the vector
$\editvec_\dataidx$. Second, if $\state$ is distant from all states $\state_\dataidx$ in the trace 
data, $\kernelvec(\state)$ is approximately a zero vector, such that the hint recommended by the 
GPR policy degrades to zero. In other words, if the student's solution is dissimilar to 
everything we have seen before, the GPR hint policy cannot provide feedback. While this limits 
coverage, it also implies that the GPR hint policy automatically detects where its hints may not be 
useful anymore, namely in case of truly novel strategies for which new example data are required.

As an example for the application of the GPR hint policy, consider the string edit 
distance example shown in Figure~\ref{fig:chf_edit_distance_space}. Note that the string edit 
distances are: $\dist_{\edits, \editCost}(\state, \state_1) = \dist_{\edits, \editCost}(\state, 
\state_2) = 1$ and $\dist_{\edits, \editCost}(\state_1, \state_2) = \dist_{\edits, 
\editCost}(\state_2, \state_1) = 1$. For the length scale $\bandwidth = 1$ and a noise variance 
$\nDevNoise^2 = 0$ we obtain 
\begin{equation*}
\kernelvec(\state) = (\frac{1}{\sqrt{e}}, \frac{1}{\sqrt{e}}),
\quad \kernelmat = \begin{pmatrix} 1 & \frac{1}{\sqrt{e}} \\ 
\frac{1}{\sqrt{e}} & 1 \end{pmatrix} \text{, and} \quad
\vec \gpcoeff(\state) = 
\kernelvec(\state) \cdot \kernelmat^{-1} \approx (0.3775, 0.3775).
\end{equation*}
Thus, the recommended edit, shown as orange arrow, is $\pol_\text{GPR}(\state) \approx
0.3775 \cdot \editvec_1 + 0.3775 \cdot \editvec_2$. For the Snap example, the result of the GPR
hint policy is shown in orange in Figure~\ref{fig:snap_hints}. As can be seen, the GPR policy is 
able to return a hint that represents the local trend in the data, thereby improving 
upon both the one-nearest neighbor (1-NN) and the Nadaraya-Watson regression (NWR) policy.

Via the hint policy $\pol_\text{GPR}$, we can now recommend edits in the edit distance space which 
are optimal according to a Gaussian Process model. However, this edit has the form of a 
coefficient vector $\vec \gpcoeff(\state)$, which is not directly interpretable to a student. 
So our last challenge is to derive a viable hint from our prediction in the edit distance space. 
More precisely, we wish to obtain an edit in our original edit set $\edits$ which corresponds to 
the 
recommended edit in the edit distance space.

\subsection{Edit pre-image problems}

The problem of finding an unknown original object which maps to a known point in an embedding space 
is called a \emph{pre-image problem} \cite{Bakir2003}, so the problem of finding the edit which best
corresponds to a recommended edit in the edit distance space can be described as an \emph{edit
pre-image problem}. We want to emphasize here that such pre-image problems are typically hard to
solve \cite{Bakir2003}, and, to our knowledge, no approach exists to date which addresses edit 
pre-image problems. We also note a connection to the state reification \cite{Rivers2014} because
the mapping to the edit distance space $\kernelfeat$ can be seen as a canonicalization, and we now
try to align the edit returned by a hint policy in the edit distance space with the student's
original state. In this section, we provide an approximate solution to the edit pre-image problem.

First, following Bakir et al., we re-frame our edit pre-image problem as a minimization problem:
Starting from the student's current state $\state$, we try to find an edit $\edit$ which brings us 
as close as possible to the recommended state of the Gaussian Process regression (GPR) hint policy
in the edit distance space.
\begin{equation}
\min_{\edit \in \edits}
\lVert \kernelfeat\big(\edit(\state)\big) - \big(\kernelfeat(\state) + \pol_\text{GPR}(\state)\big) 
\rVert^2
\label{eq:edit_pre_image}
\end{equation}

This optimization problem is infeasible because it requires us to estimate the effect of an 
edit $\edit$ to the student's state $\state$ after mapping this state to the edit distance space.
Fortunately, our established edit distance theory lets us replace this minimization problem with
a simpler form.

\begin{thm} \label{thm:pre_image}
Let $\vec \gpcoeff$ be the weights applied by GPR, that is,
$\vec \gpcoeff = \kernelvec(\state) \cdot (\kernelmat + \nDevNoise^2 \cdot \eye^\datalim)^{-1}$.
Then the maximization problem in \ref{eq:edit_pre_image} can be re-written as:
\begin{equation}
\min_{\edit \in \edits} \dist_{\edits, \editCost}(\edit(\state), \state)^2 +
\sum_{\dataidx=1}^\datalim \coeff_\dataidx \cdot
	\dist_{\edits, \editCost}(\edit(\state), \state_\dataidx)^2
\label{eq:edit_pre_image_revised}
\end{equation}
where $\coeff_\dataidx = - \gpcoeff_\dataidx$ if $\state_\dataidx$ is the start point of a trace,
$\coeff_\dataidx = \gpcoeff_{\dataidx-1} - \gpcoeff_\dataidx$ if $\state_\dataidx$ is an 
intermediate element of a trace, and $\coeff_\dataidx = \gpcoeff_{\dataidx - 1}$ if 
$\state_\dataidx$ is the end point of a trace.
\begin{proof}
In a first step, we note that we can re-write:
\begin{equation}
\kernelfeat(\state) + \pol_\text{GPR}(\state)
= \kernelfeat(\state) + \sum_{\dataidx = 1}^\datalim \coeff_\dataidx \cdot 
\kernelfeat(\state_\dataidx)
\end{equation}
According to Theorems 3 and 4 in the paper by Paaßen 
et al., we can now re-write the distance $\lVert \kernelfeat\big(\edit(\state)\big) - 
\big(\kernelfeat(\state) + \pol_\text{GPR}(\state)\big) 
\rVert^2$ as
\begin{equation}
\lVert \kernelfeat(\edit(\state)) - \kernelfeat(\state) \rVert^2 +
\sum_{\dataidx=1}^\datalim \coeff_\dataidx \cdot \lVert \kernelfeat\big(\edit(\state)\big) - 
\kernelfeat\big(\state_\dataidx\big) \rVert^2 - Z
\end{equation}
where $Z$ is a constant that does not depend on $\edit$ \cite{Paassen2017NPL}. Due to
Theorem~\ref{thm:edit_distance_space} we know that the Euclidean distance in the edit distance
space corresponds exactly to the edit distance, which concludes the proof.
\end{proof}
\end{thm}

This form of the minimization problem in Equation~\ref{eq:edit_pre_image_revised} has multiple key
advantages. First, it does not require us to compute the vectorial embedding for any 
state anymore. Instead, we can infer the optimal edit sequence solely based on the edit distance 
$\dist_{\edits, \editCost}(\edit(\state), \state)$, as well as the edit distances $\dist_{\edits, 
\editCost}(\edit(\state), \state_\dataidx)$. Second, our revised form of the problem provides a 
useful re-interpretation. We need to find an edit $\edit$, such that the resulting state stays 
close to the original state $\state$, gets closer to states $\state_\dataidx$ for which 
$\coeff_\dataidx$ is positive, and gets further away from state $\state_\dataidx$ for which 
$\coeff_\dataidx$ is negative. Note that this re-interpretation is consistent with the 
criterion of \citeN{Rivers2014} that a next state should stay close to the student's 
current state. Finally, the re-formulation shrinks our search space, 
because we only have to consider edits which bring us closer to states $\state_\dataidx$ with 
positive coefficients $\coeff_\dataidx$. We can extract such edits from the shortest edit sequences 
between $\state$ and states $\state_\dataidx$ with positive coefficients $\coeff_\dataidx$.
For all these possible edits we can evaluate the error in Equation~\ref{eq:edit_pre_image_revised}
and select the edit with the lowest error.

Consider the example illustrated in Figure~\ref{fig:chf_pre_image}. Recall that the coefficients
$\coeff$ resulting from the GPR hint policy are 
$\coeff_{\state_1} = \coeff_{\state_2} \approx -0.3775$ and
$\coeff_{\outstate_1} = \coeff_{\outstate_1} \approx +0.3775$. So we need to find an edit which
brings us closer to $\outstate_1 = \text{aac}$ and $\outstate_2 = \text{bbc}$ but further away from
$\state_1 = \text{a}$ and $\state_2 = \text{b}$. The cheapest edit sequence between
$\state$ and $\outstate_1$ is $\rep_{2,\text{a}}, \ins_{3, \text{c}}$, and the cheapest
edit sequence between $\state$ and $\outstate_2$ is $\rep_{1,\text{b}}, \ins_{3, \text{c}}$.
Therefore, we need to consider the edits $\rep_{2,\text{a}}$, $\ins_{3, \text{c}}$, 
and $\rep_{1,\text{b}}$. The resulting states of these edits would be aa, abc, and bb. Amongst 
these options, abc minimizes our error because it is closer to both aac and bbc, further away from 
both a and b, and stays close to ab. Therefore, we would recommend $\ins_{3, \text{c}}$ as hint.

In practical examples, this approach would be limited by the number of edits to be considered. For 
many training data points with positive coefficients and long shortest edit sequences, this number 
can become unreasonable. One way to limit the number of edits is to incorporate more of the 
criteria suggested by \citeN{Rivers2014} and consider only edits which result in syntactically 
correct states, result in programs which fulfill at least as many test cases or get us closer to 
a correct solution. In addition, we propose to limit the search space to a 
reasonable size by using fewer coefficients to represent the recommended state. More precisely, we 
are looking for a coefficient vector $\tilde \coeff$ which is nonzero for at most $\sparselim$ 
training states. Further, we propose to use only those states to represent the recommended state 
which are between the student's current state $\state$ and the next correct solution $\state^*$. 
This is consistent with the criteria of \citeN{Rivers2014} that the recommended state should both 
be close to a correct solution and to the student's current state. 
Thus, we look for a coefficient vector $\tilde \coeff$, such that $\tilde \coeff_\dataidx$ is 
nonzero only if $\dist(\state_\dataidx, \state) \leq \dist(\state, \state^*)$ and 
$\dist(\state_\dataidx, \state^*) \leq \dist(\state, \state^*)$, such that at most $\sparselim$ 
entries are nonzero, such that the sum over all entries of $\tilde \coeff$ is $1$, and such that 
the state represented by $\tilde \coeff$ is as close as possible to the state represented by $\vec 
\coeff$. While this is an NP-hard problem, multiple simple heuristics exist which have been 
summarized by \citeN{Hofmann2014}. In our experiments, we apply both kernelized 
orthogonal matching pursuit and an approximation via the largest entries of $\vec \coeff$ and use 
whatever approximation is closer to the actual recommended state.

\begin{figure}
\begin{center}
\begin{tikzpicture}[scale=1.5]
\begin{scope}[shift={(-1, 0)}]
\node[defaultcolor]   (x1) at (140:0.9) {a};
\node[defaultcolor]   (x2) at (220:0.9) {b};
\end{scope}

\node[user1color]   (x) at (-1, 0) {ab};

\node[defaultcolor] (aa) at (120:1) {aa};
\node[defaultcolor] (y1) at ( 60:1) {aac};

\node[defaultcolor] (bb) at (240:1) {bb};
\node[defaultcolor] (y2) at (300:1) {bbc};


\node[hint3color] (g)  at (2, 0) {abcd};

\draw[draw=plum3, semithick, dashed] (2,0) arc (0:60:2);
\draw[draw=plum3, semithick, dashed] (2,0) arc (0:-60:2);
\draw[draw=plum3, semithick, dashed] (-1,0) arc (180:270:3 and 1.2);
\draw[draw=plum3, semithick, dashed] (-1,0) arc (180:90:3 and 1.2);

\begin{scope}[node distance=0.5cm]
\node[hint1color] [above of=x1] {-0.3775};
\node[hint1color] [below of=x2] {-0.3775};
\node[hint1color] [above of=y1] {+0.3775};
\node[hint1color] [below of=y2] {+0.3775};
\node[hint1color] [right of=x]  {1};
\end{scope}

\node[diamond, hint1, semithick] at (0.653, 0) {};

\begin{scope}[node distance=1cm]
\node[hint2color] [above of=y1] {+0.3043};
\node[hint2color] [below of=y2] {+0.3043};
\end{scope}
\node[hint2color] [right of=g, node distance=1.5cm]  {+0.3914};

\node[diamond, hint2, semithick] at (1.087, 0) {};

\end{tikzpicture}
\end{center}
\caption{An illustration of the sparse representation of the recommended state for
the string example of Figure~\ref{fig:chf}. The student's current state
is the string $\state = \text{ab}$ (shown in red), the closest correct
solution is the string $\state^* = \text{abcd}$ (shown in green).
The coefficients $\coeff_\dataidx$ and the represented state $\kernelfeat(\state) + \pol_\text{GPR}(\state)$
returned by the Gaussian Process Regression (GPR) policy are shown in 
orange. The sparse coefficients and the corresponding represented state are drawn in purple. The 
constraints $\dist(\state_\dataidx, \state) \leq \dist(\state, \state^*)$ and 
$\dist(\state_\dataidx, \state^*) \leq \dist(\state, \state^*)$ are illustrated by dashed purple 
lines.}
\label{fig:sparsity}
\end{figure}
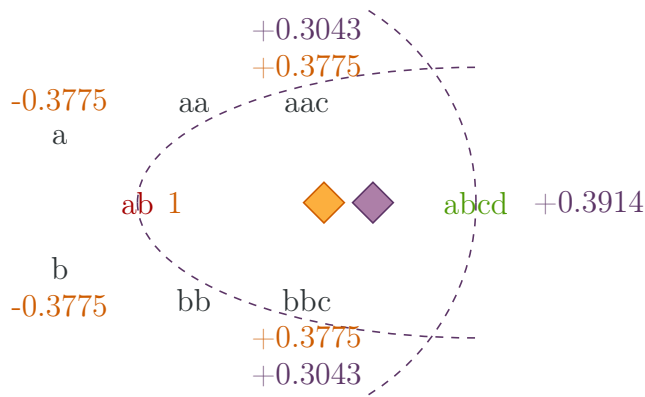

Consider the example illustrated in Figure~\ref{fig:sparsity}. Here, the original coefficients 
$\coeff$ returned by the GPR hint policy are shown in orange and represent the state shown as an
orange diamond. Now, assume that the student's current state is the string \enquote{ab} and 
the closest correct solution is the string \enquote{abcd}. In that case, only the 
strings \enquote{ab,} \enquote{aac,} \enquote{bbc,} and \enquote{abcd} fulfill the 
constraints $\dist(\state_\dataidx, \state) \leq \dist(\state, \state^*)$ and 
$\dist(\state_\dataidx, \state^*) \leq \dist(\state, \state^*)$ (indicated by dashed purple 
lines). If we now try to represent the recommended state by using only $3$ of those 
four strings, this results in a representation via the strings \enquote{aac,} \enquote{bbc,} and 
\enquote{abcd} with roughly equal coefficients, resulting in a represented state (shown 
in purple) close to the original hint. The selected hint, in this case, would still be
$\ins_{3, \text{c}}$.

\subsection{Summary}

To conclude our description of the Continuous Hint Factory (CHF) we provide a short summary
of all steps involved in the CHF hint policy. First, we need to perform the following
preparation steps:

\begin{enumerate}
\item Collect trace data from successful students.
\item Remove all intermediate states in the traces which do not get closer to the goal.
\item Compute the canonic forms of the trace data and their pairwise edit distances.
\item Perform eigenvalue correction on the pairwise edit distances.
\item Compute the pairwise radial basis kernel values $\kernelmat$.
The length scale parameter $\bandwidth$, as well as the noise parameter $\nDevNoise$, can be selected 
such that the predictive accuracy of the GPR model on unseen evaluation data is as high as possible.
\end{enumerate}

Now, assume that a new student is in state $\state$ and requests help. In that case, the following 
steps need to be performed.

\begin{enumerate}
\item Compute the canonic form of $\state$ and the edit distance of this canonic form to all 
canonic forms in the trace data before.
\item Extend the eigenvalue correction to the new distances.
\item Compute the radial basis kernel values $\kernelvec(\state)$ based on these corrected 
distances.
\item Compute the coefficients $\coeff$ of the GPR hint policy via the formulas in 
Theorem~\ref{thm:pre_image}.
\item Optionally, sparsify these coefficients via one of the techniques of 
\citeN{Hofmann2014}.
\item Compute the cheapest edit scripts between $\state$ and all training states $\state_\dataidx$ 
for which $\coeff_\dataidx$ is positive.
\item Subselect edits $\edit$ from these edit scripts which result in states $\edit(\state)$ that 
conform to further criteria, e.g., unit test fulfillment, or syntactic correctness \cite{Rivers2014}.
\item Compute the error term in Equation~\ref{eq:edit_pre_image_revised} for all remaining edits.
\item Select the edit with the lowest error as hint.
\end{enumerate}

This concludes our description of the CHF. In the next section, we evaluate the CHF approach 
experimentally.

\section{Experiments}
\label{sec:experiments}

We consider two datasets for our analysis. First, a dataset collected in an introductory 
undergraduate computing course for non-computer science majors during the Fall of 2015
at a research university in the south-eastern United States. The course had approximately 80
students, split among six lab sections. The first half of the course focused on learning the
Snap\footnote{\url{http://snap.berkeley.edu}} programming language through a curriculum based on
the \emph{Beauty and Joy of Computing} \cite{Garcia2015}. Here, we focus on the \enquote{Guessing Game} 
task, which had the following description: \enquote{The computer chooses a random number between 1 
and 10 and continuously asks the user to guess the number until they guess correctly.} Students did 
not receive specific instructions regarding the form of the program. An example solution for the 
task is presented in Figure~\ref{fig:snap}. Students worked on this assignment during class for
approximately one hour, with a teaching assistant available to assist them and the option of
working in pairs. The class was conducted as normal, and the students were not informed that data
was being collected. The state of the student's program was recorded after every edit. Students who
did not correctly select the assignment they were working on were excluded from the analysis. The
dataset consists of $52$ traces with $8669$ states overall.

Each of the final states was graded by two independent graders. The graders used a rubric
consisting of nine assignment objectives and marked whether each state successfully or
unsuccessfully completed each objective. The graders had an initial agreement of 94.5\%, with
Cohen's $\kappa = 0.544$. After clarifying objective criteria, each grader independently regraded
each state where there was disagreement, reaching an agreement of $98.1\%$, with Cohen's $\kappa
= 0.856$. Any remaining disagreements were discussed to create final grades for each assignment. As 
our aim is to predict what \emph{capable} students would do, we kept only traces which successfully 
completed at least eight of the nine objectives. This left $47$ traces with $7864$ states.

\tikzstyle{umlstart}=[circle, fill=black, draw=none, minimum size=0.3cm]
\tikzstyle{umlaction}=[rectangle, fill=none, draw=black, thick, rounded corners]
\tikzstyle{umldecision}=[diamond, fill=none, draw=black, thick, minimum size=0.8cm, sharp corners]
\tikzstyle{umlmerge}=[umldecision]
\tikzstyle{umlfinish}=[circle, fill=black, draw=black, double=white, double distance=0.05cm, 
minimum size=0.35cm]
\tikzstyle{umledge}=[->,shorten >=1pt,shorten <=1pt, thick, >=stealth']

\begin{figure}
\begin{center}
\begin{tikzpicture}
\node[umlstart] (start) at (0,0) {};
\node [right of=start, node distance=1cm] {0};
\node[umlaction, align=center] (initx) at (0,-1.25) {Let $x$ be the first input.\\ Let $m = 
|x|$.};
\node [right of=initx, node distance=3cm] {1};
\node[umlaction, align=center] (inity) at (0,-3) {Let $y$ be the second input.\\ Let $n = 
|y|$.};
\node [right of=inity, node distance=3cm] {2};
\node[umldecision] (paddec) at (0,-4.5) {};
\node [above right of=paddec, node distance=0.75cm] {3};
\node[umlaction, align=center] (padx) at (-2,-5.75)  {Pad $x$ with zeros\\to length $n$.};
\node [left of=padx, node distance=2.5cm] {4};
\node[umlaction] (padx2) at (-2,-7.25) {$m \gets n$.};
\node [left of=padx2, node distance=1.5cm] {5};
\node[umlaction, align=center] (pady) at (2,-5.75) {Pad $y$ with zeros\\to length $m$.};
\node [right of=pady, node distance=2.5cm] {6};
\node[umlmerge] (padmerge) at (0,-8) {};
\node [above of=padmerge, node distance=0.75cm] {7};
\node[umlaction, align=center] (initz) at (0,-9.25) {Initialize $z$ as binary number with 
$m$ digits.};
\node [right of=initz, node distance=4.5cm] {8};
\node[umlaction, align=center] (initc) at (0,-10.5) {Initialize $c \gets 0$.};
\node [right of=initc, node distance=2cm] {9};
\node[umlaction, align=center] (initi) at (0,-11.75) {Initialize $i \gets m$.};
\node [right of=initi, node distance=2cm] {10};
\node[umldecision] (dec) at (0,-13) {};
\node [above left of=dec, node distance=0.75cm]   {11};
\node[umldecision] (bindec) at (0,-14.25) {};
\node [above right of=bindec, node distance=0.75cm] {12};
\node[umlaction] (z0) at (-5,-16.25) {$z_i \gets 0$.};
\node [left of=z0, node distance=1.3cm] {14};
\node[umlaction] (z1) at (-1.75,-15.5) {$z_i \gets 1$.};
\node [left of=z1, node distance=1.3cm] {15};
\node[umlaction] (c1) at (-1.75,-16.75) {$c \gets 0$.};
\node [left of=c1, node distance=1.3cm] {16};
\node[umlaction] (z2) at (+1.75,-15.5) {$z_i \gets 0$.};
\node [right of=z2, node distance=1.3cm] {17};
\node[umlaction] (c2) at (+1.75,-16.75) {$c \gets 1$.};
\node [right of=c2, node distance=1.3cm] {18};
\node[umlaction] (z3) at (+5,-16.25) {$z_i \gets 1$.};
\node [right of=z3, node distance=1.3cm] {19};
\node[umlmerge] (binmerge) at (0,-17.75) {};
\node [below right of=binmerge, node distance=0.75cm] {20};
\node[umlaction] (idec) at (0,-19) {$i \gets i-1$.};
\node [right of=idec, node distance=1.5cm] {21};
\node[umlaction, align=center] (return) at (3,-13) {return $z$.};
\node [above of=return, node distance=0.5cm] {22};
\node[umlfinish] (fin) at (5,-13) {};
\node [right of=fin, node distance=1cm] {23};

\path[umledge]%
(start)  edge (initx)
(initx)  edge (inity)
(inity)  edge (paddec)
(paddec) edge node[above left] {$m \leq n$} (padx) edge node[above right] {$n < m$} (pady)
(padx)   edge (padx2)
(padx2)  edge[out=270, in=180] (padmerge)
(pady)   edge[out=270, in=0] (padmerge)
(padmerge) edge (initz)
(initz)  edge (initc)
(initc)  edge (initi)
(initi)  edge (dec)
(dec)    edge node[left] {$0 < i$} (bindec)
(bindec) edge [out=180,in=90] node[above] {$x_i + y_i + c = 0$} (z0)
(bindec) edge node[left] {$x_i + y_i + c = 1$} (z1)
(bindec) edge node[right] {$x_i + y_i + c = 2$} (z2)
(bindec) edge[out=0,in=90] node[above] {$x_i + y_i + c = 3$} (z3)
(z0)     edge[out=270,in=180] (binmerge)
(z1)     edge (c1)
(c1)     edge (binmerge)
(z2)     edge (c2)
(c2)     edge (binmerge)
(z3)     edge[out=270,in=0] (binmerge)
(binmerge) edge (idec)
(idec)   edge [out=180,in=180, looseness=3.5] (dec)
(dec)    edge node[above] {$i = 0$} (return)
(return) edge (fin);
\end{tikzpicture}
\end{center}
\caption{A correct example solution for the UML binary adder task. Numbers indicate the order in 
which nodes have been added to the UML diagram.}
\label{fig:uml_example}
\end{figure}
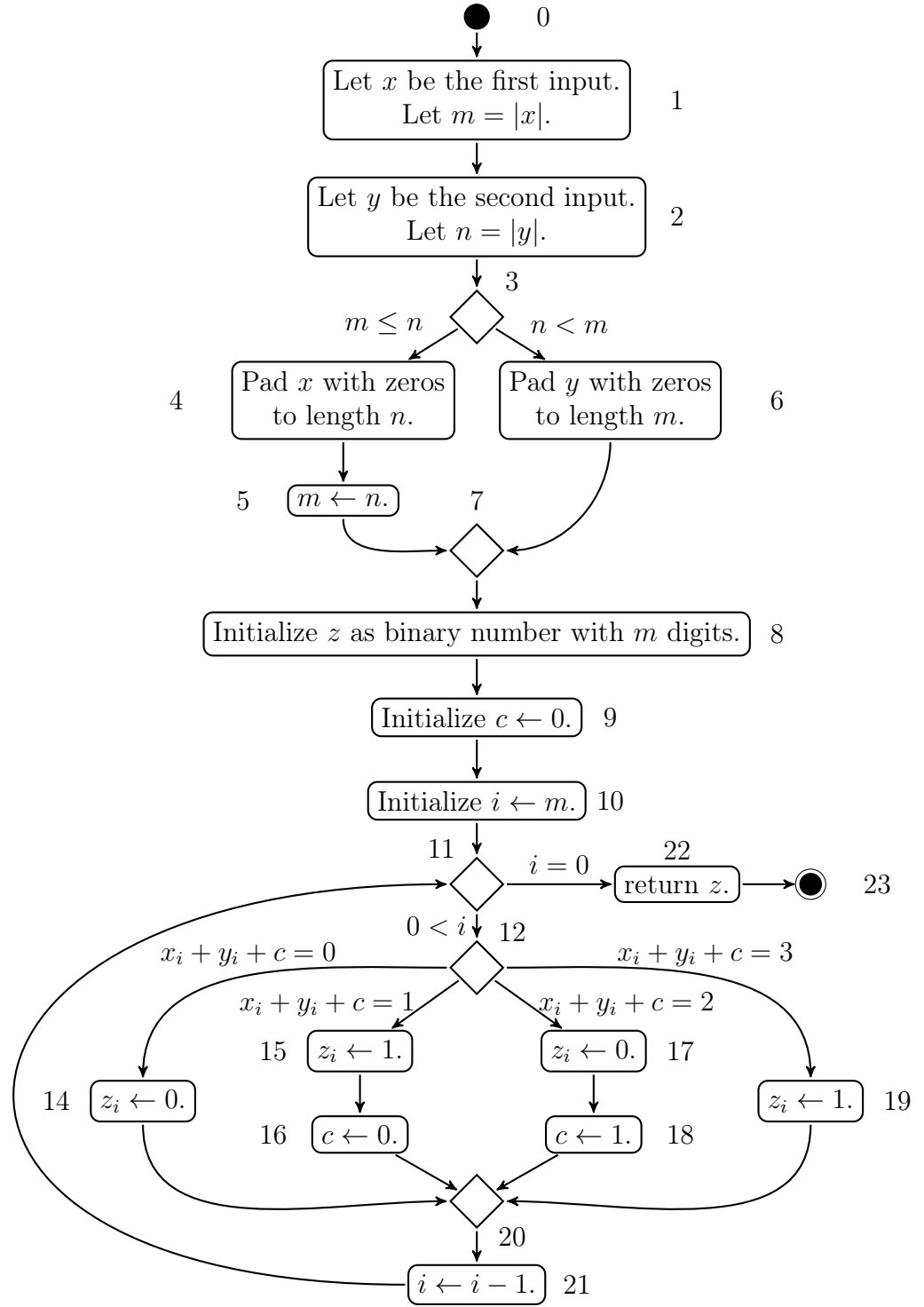

As a second dataset, we utilize data collected in an introductory programming course for computer 
scientists at a German university in 2012. The students were asked to draw a UML 
activity diagram which described the process of adding two binary numbers. An example solution 
is shown in Figure~\ref{fig:uml_example}. From the available student data, we extracted six 
typical strategies and created two correct traces and one erroneous trace for each strategy. 
Overall, the correct traces contained $364$ states and the erroneous traces $115$ states. We 
presented each state in the erroneous traces to three graders who independently were asked to 
suggest all possible edit hints that could be given to a student in the particular situation, 
taking past states into account. We also instructed the tutors to provide an estimate of hint 
quality in the interval $[0, 1]$ for each of their hints, taking into account the following 
criteria:
1) Does the hint follow the strategy of the student?
2) Does the hint conform to the student's current focus of attention or does it address a different
part of the state?
3) Is the hint effective in addressing the problems in the student's state?
4) Is the hint effective in guiding the student toward a solution?
In a second meeting, all tutors met to add ratings for the hints of the respective other tutors and 
to discuss discrepancies in the ratings.
If after discussion at least one expert rated a hint with a grade below $0.5$, the hint was excluded
from the set. $1053$ hints remained after this process. The average inter-rater correlation via 
Pearson's $r$ was $r = 0.588$, indicating moderate agreement. This dataset is available online 
under the DOI \href{http://doi.org/10.4119/unibi/2913083}{10.4119/unibi/2913083}.

We represent the states of both datasets as trees. In the UML dataset, we removed 
back-references, such as the arrow from node 20 to node 11 in Figure~\ref{fig:uml_example},
to obtain a tree structure. We labeled all tree nodes with the name of the respective 
syntactic construct (e.g., \enquote{doIf,} \enquote{var,} and \enquote{literal}). In case of the UML 
dataset, we also added the text of the respective node to the label, for example, \enquote{return 
$z$} for node 21 in Figure~\ref{fig:uml_example}. In both datasets, we canonicalized the
trees by normalizing variable names and literals, normalizing the order of binary relations, and
removing non-executable code, as recommended by \citeN{Rivers2012}.

As an edit distance, we employ the tree edit distance of \citeN{Zhang1989} as 
implemented in the \emph{TCS Alignment Toolbox} \cite{Paassen2015EDM}. For the Snap dataset we 
use a uniform edit cost of $1$ for deletions, insertions, and replacements. For the UML dataset,
we define deletion and insertion costs as $1$, replacement costs between unequal node types 
as infinite, and replacement costs between action nodes (displayed as ellipses in 
Figure~\ref{fig:uml_example}) as the string edit distance between the node text, normalized to the 
interval $[0, 1]$. We post-process the tree edit distances via clip eigenvalue correction 
\cite{Gisbrecht2015}. Based on the tree edit distance, we excluded states which did not get closer 
to the final state in the respective trace, which left $1005$ states. Of these states, $812$ were 
unique, and of these unique states, $94.09\%$ were visited only once. Similarly, $215$ of the $354$ 
training states in the UML dataset were unique, and of these unique states, $82.79\%$ were visited 
only once. These numbers indicate that meaningful frequency information is only available for
very few training states, which is consistent with the findings reported by \citeN{Price2015}
on similar data from an open-ended Snap programming task.

To evaluate the utility of the CHF fairly, we need to compare to existing reference hint policies.
Due to the lack of meaningful frequency information in our data, however, we can neither apply
the Hint Factory \cite{Barnes2008} nor the Piech policy \cite{Piech2015}. Furthermore, to keep the approach 
generic, we do not use task-specific syntactic or unit test information for our experiments, which 
rules out the policy of \citeN{Lazar2014}. There remain the policy of \citeN{Gross2015AIED}, which 
uses the successor of the next state in the trace data to construct a hint, the 
policy of \citeN{Zimmerman2015}, which uses the closest correct solution to construct a 
hint, and the policy of \citeN{Rivers2015}, which also uses the closest
correct solution. The Zimmerman policy and the Rivers policy mainly differ in 
\emph{how} hints are constructed from the closest correct solution. However, given that we 
use neither frequency nor syntactic or semantic correctness information, and consider only
single edits instead of edit combinations, both policies become very similar, such that we only
consider the Gross policy and the Zimmerman policy in this case.

We implemented all hint policies in MATLAB\textsuperscript{\textregistered} \footnote{Our implementation of the GPR hint policy 
is available under the DOI \href{http://doi.org/10.4119/unibi/2913104}{10.4119/unibi/2913104}}.
To optimize the kernel length scale $\bandwidth$ and the noise standard deviation $\nDevNoise$ of 
the Gaussian Process model, we employ a random hyper-parameter search with $10$ repeats as 
recommended by \citeN{Bergstra2012}. We set the maximum number of training 
states to represent the hint of the CHF policy to $\sparselim = 11$.

In our experiments, we investigate two research questions, which we will cover in turn. We 
evaluate statistical significance using a paired Wilcoxon sign-rank test. Further, we apply a 
Bonferroni correction to avoid type I errors due to multiple tests.

\paragraph{RQ1:} How well does the Gaussian Process model capture the behavior of capable students,
that is, can the Gaussian Process predict what a capable student would do?

To investigate RQ1, we consider two measures of predictive accuracy. First, we measure the distance
between the predicted next state of the Gaussian Process model and the actual next state of the 
respective student (next-step error). Second, we measure the distance between the predicted next state 
and the \emph{final} state of the respective student (final-step error). We square these distances,
average them over a trace, and then compute the root, resulting in a
root mean square error (RMSE), which estimates the standard deviation of the error distribution of
the model and is a well-established measure for model evaluation \cite{Chai2014}.
We evaluate the next-step error and the final-step error in a leave-one-out crossvalidation
over the traces, which means that in each fold we use all but one trace as training data for
the prediction and the remaining trace to evaluate the model.

Note that RQ1 is only concerned with the prediction module of each hint policy, that is, the
reference state based on which edits are generated, not the edits which are used as hints.
As such, we do not directly compare with the Gross or Zimmerman policy but with the reference
states they would use, namely the successor of the closest next solution (Successor-of-closest),
and the closest correct solution (Closest-correct) respectively. Given the nature of
these references, we would expect that the Successor-of-closest prediction would perform well
in the next-step error but badly in the final-step error and that the Closest-correct
prediction would perform badly in terms of the next-step error but good in terms of the final-step
error.

\begin{table}
\caption{Mean RMSE $\pm$ standard deviation in predicting the next step and 
the final step of capable students for both the Snap dataset, 
as well as the UML dataset. The first column lists the different prediction schemes.
Lower values are better, and a value of $0$ is ideal.}
\begin{center}
\begin{tabular}{lcccc}
\cmidrule{1-5}
            & \multicolumn{2}{c}{Snap} & \multicolumn{2}{c}{UML} \\
            \cmidrule(lr){2-3} \cmidrule(lr){4-5}
Prediction scheme    & Next & Final & Next & Final  \\
\cmidrule(lr) {1-1} \cmidrule(lr) {2-2} \cmidrule(lr) {3-3} \cmidrule(lr) {4-4} \cmidrule(lr) {5-5}
Do nothing           & $ 17.5\pm 3.89$ & $ 39.3\pm 9.36$ & $ ~~5.27\pm 0.53$ & $ 28.9\pm 5.28$ \\
Successor-of-closest & $ 23.7\pm 5.39$ & $ 39.1\pm 9.49$ & $ ~~7.89\pm 3.50$ & $ 29.1\pm 6.00$ \\
Closest-correct      & $ 26.7\pm 5.46$ & $ 43.0\pm 8.60$ & $ 25.50\pm 1.23$ & $ 19.9\pm 8.42$ \\
Gaussian Process     & $ 16.6\pm 4.09$ & $ 37.8\pm 9.15$ & $ ~~3.18\pm 1.66$ & $ 27.8\pm 5.32$ \\
\cmidrule{1-5}
\end{tabular}
\end{center}
\label{tab:results_rq1}
\end{table}

Table~\ref{tab:results_rq1} shows the RMSE averaged over the crossvalidation folds ($\pm$ standard deviation)
for both datasets where each column lists one error measure for all prediction schemes\footnote{Note that the RMSE cannot
be interpreted directly as the average number of edits between the predicted next state and the
gold standard because the RMSE assigns higher weight to larger deviations due to the square
\cite{Chai2014}. Further, in this particular evaluation, but not for RQ2, Eigenvalue correction
distorts the edit distances to become larger.}.
As an additional reference, we provide the error for the trivial prediction of staying in the
same state, that is, $\pol(\state) = \state$.
Statistical analysis reveals that the Gaussian Process is significantly better in predicting the
next state compared to all other baselines for both datasets ($p < .01$). Further, the Gaussian
Process is significantly better in predicting the final state compared to the \enquote{Do nothing}
and the Successor-of-closest prediction for both datasets ($p < .01$), and better than the
Closest-correct prediction for the Snap dataset ($p < .001$).
Interestingly, the Successor-of-closest prediction does not perform better in predicting the actual
next state of a student compared to staying in the same state, indicating that students in both data
sets do not necessarily move along the same states, even though their directions may be consistent,
which is consistent with the embedding in Figure~\ref{fig:snap_embedding}.
Furthermore, we note that, counter to our expectations, the Closest-correct prediction has 
a higher final-step error on the Snap dataset than any other prediction scheme, which indicates
that, on average, the closest correct solution of other students is far away from the student's
actual final solution in this dataset. This effect is likely explained by the high strategic
variability in an open-ended programming task such as the guessing game task. For such tasks,
we expect that the averaging approach of the Gaussian Process to be particularly helpful, because
the general trends in the datasets may be more akin to the student's actual plans than a single
closest correct solution.
Conversely, the UML dataset features less strategic variability, and the closest correct solution of
another student is still close to the final state of the student for which the prediction is made,
which is reflected in significantly better predictions of the Closest-correct prediction compared to
all other prediction schemes ($p<10^{-3}$).
Overall, we can conclude that the Gaussian Process is more accurate in predicting the next state
of students compared to other baselines on our example datasets and that this is especially the
case for the Snap dataset, which is characterized by high strategic variability.

\paragraph{RQ2:} Do the hints of the Continuous Hint Factory correspond to the hints of human 
tutors?

To investigate RQ2, we require a reference measure of hint quality, which is provided by the 
quality judgments of human tutors in the UML dataset. In particular, we iterate over every state in 
the erroneous traces of the UML dataset and generate a hint with each hint policy, using all 
correct traces as training data. If multiple edits achieve the lowest error rank, we resolve
ties by selecting the edit as hint which is closest to the root of the tree. If the recommended
hint of the policy matches at least one tutor hint, we assign the average quality rating of the
human tutors for that hint. Otherwise we set the rating to $0$. This is similar to the evaluation
scheme suggested by \citeN{Price2017EDM}.
We report five evaluation measures, namely the median and mean hint quality, the fraction 
of hints with a quality $>0$, the distance between the policy hint and the closest human tutor 
hint in terms of RMSE, and the fraction of states for which a hint could be generated.
In addition to the Gross and the Zimmerman policy, we also compare to a random policy, which 
selects a random reference state from the training state and recommends an edit on the 
shortest path towards that state as hint. Finally, we also provide the best-rated tutor hint 
as the gold standard.

\begin{table}
\caption{The hint evaluation measures for all hint policies on the UML dataset.
Mean hint quality and mean ambiguity are reported with standard deviation. For 
all measures except the RMSE, higher numbers are better with a value of $1$ and $100\%$ 
respectively being ideal.}
\begin{center}
\begin{tabular}{lccccc}
\cmidrule{1-6}
                     & \multicolumn{3}{c}{Hint quality}          & RMSE & Hintable \\
\cmidrule(lr){2-4} \cmidrule(lr){5-5} \cmidrule(lr){6-6}
Hint policy          & Median  & Mean    & $>0$      &      &           \\
\cmidrule(lr){1-1} \cmidrule(lr){2-2} \cmidrule(lr){3-3} \cmidrule(lr){4-4}
Random               & $0.0$ & $0.360 \pm 0.456$ & $~~39.1\%$ & $1.42$ & $~~83.5\%$ \\
Tutor                & $1.0$ & $0.994 \pm 0.021$ & $100.0\%$  & $0.00$ & $100.0\%$ \\
Gross                & $0.8$ & $0.569 \pm 0.465$ & $~~60.9\%$ & $1.42$ & $100.0\%$ \\
Zimmerman            & $0.8$ & $0.557 \pm 0.431$ & $~~64.3\%$ & $1.48$ & $100.0\%$ \\
CHF                  & $0.9$ & $0.590 \pm 0.471$ & $~~61.7\%$ & $1.36$ & $~~97.4\%$ \\
\cmidrule{1-6}
\end{tabular}
\end{center}
\label{tab:results_rq3}
\end{table}

\begin{figure}
\begin{center}
\hspace{3cm}
\begin{tikzpicture}[node distance=1.5cm]
\node[umlstart] (start) at (0,0) {};
\node [left of=start, node distance=1cm] {0};
\node[umlaction, align=center] (initx) at (0,-1.25) {Let $x$ be the first input.\\ Let $m = 
|x|$.};
\node [left of=initx, node distance=3cm] {1};
\node[umlaction, align=center] (inity) at (0,-3) {Let $y$ be the second input.\\ Let $n = 
|y|$.};
\node [left of=inity, node distance=3cm] {2};
\node[umlaction, align=center] (initz) at (0,-4.75) {Initialize $z$ as binary number with 
$m$ digits.};
\node [left of=initz, node distance=4.5cm] {3};
\node[umlaction, align=center] (initc) at (0,-6) {Initialize $c \gets 0$.};
\node [left of=initc, node distance=2cm] {4};
\node[umlaction, align=center] (initi) at (0,-7.25) {Initialize $i \gets m$.};
\node [left of=initi, node distance=2cm] {5};
\node[umldecision] (dec) at (0,-8.5) {};
\node [left of=dec, node distance=1cm]   {6};
\node[umlaction, align=center] (return) at (3,-8.5) {return $z$.};
\node [above of=return, node distance=0.5cm] {7};

\path[umledge]%
(start)  edge (initx)
(initx)  edge (inity)
(inity)  edge (initz)
(initz)  edge (initc)
(initc)  edge (initi)
(initi)  edge (dec)
(dec)    edge node[above] {$i < 0$} (return);

\node[umlfinish, hint1] (chf) at (5,-8.5) {};
\node[hint1color] [right of=chf] {CHF policy};

\path[umledge, hint1, densely dashed] (return) edge (chf);


\node[umldecision, hint2] (zim) at (5,-4.75) {};
\node[hint2color] [below of=zim, node distance=1cm] {Zimmerman policy};
\path[umledge, hint2, densely dashed] (inity) edge (zim);
\end{tikzpicture}
\end{center}
\caption{An example state from the UML dataset, where the Zimmerman policy generates a hint 
(purple) which is not in the student's current focus of attention. In contrast, the 
Continuous Hint Factory generates a hint (orange), which adds upon the last added node.}
\label{fig:uml_hints}
\end{figure}
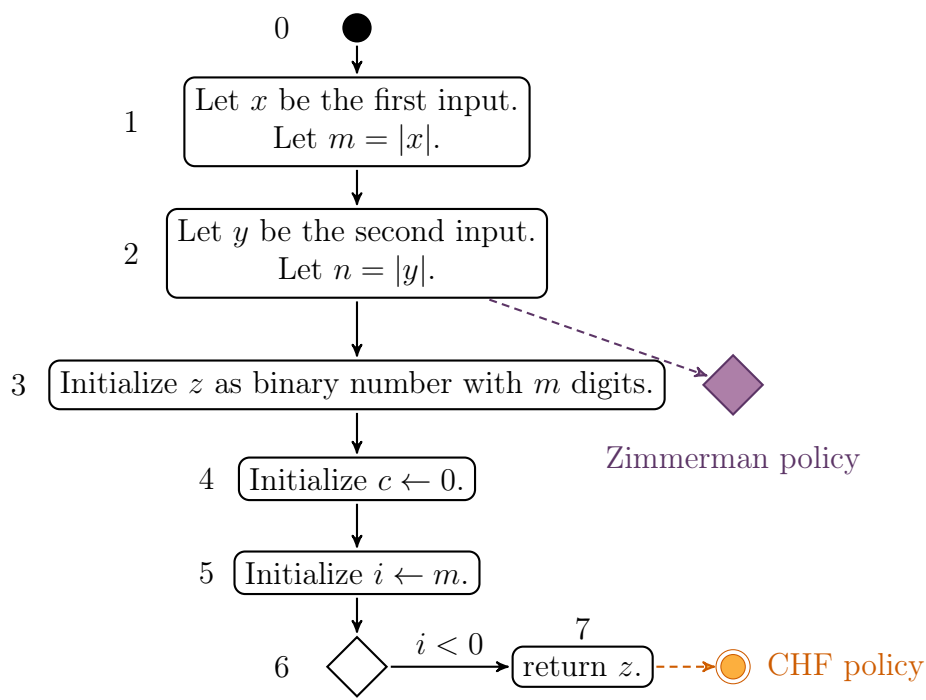

The experimental results are shown in Table~\ref{tab:results_rq3}, where each column displays one 
evaluation measure, and each row lists the results for one hint policy. Regarding hint quality, we 
observe that the CHF performs significantly better compared to a random policy ($p < .01$), and 
significantly worse compared to human tutor hints ($p < .001$), but otherwise there are no 
significant differences between the hint policies. This indicates that for simple datasets like
the UML dataset, which feature low strategic variability, single reference states are sufficient
to generate viable hints. Interestingly, though, we could also observe cases where this was not the
case. In particular, Figure~\ref{fig:uml_hints} displays a UML diagram where the Zimmerman policy
recommends appending a decision node close to the root (purple), which is outside the student's
current focus of attention because the last node the student added was the \enquote{return z} node
at the bottom of the diagram. Accordingly, the CHF recommends appending a \enquote{finish} node to
that branch (orange).

Another interesting finding is that the CHF and the Gross policy consistently achieved perfect hint 
quality for the first three steps in each trace. This is important in light of the research of
\citeN{Price2017AIED}, which indicates that students are more likely to seek help and follow hints if \emph{early} 
hints provided by the system were useful.

\section{Conclusion}
\label{sec:conclusion}

This work makes three primary contributions. First, we have provided a mathematical framework for 
edit-based hints and placed prior contributions within this framework. Second, we have introduced 
the concept of the edit distance space, which is a continuous embedding of student states such that
the edit distance corresponds to the Euclidean distance in the embedding space.
Finally, we introduced the \emph{Continuous Hint Factory} (CHF), a novel hint policy
which provides edit hints to students by choosing an edit consistent with the general trend of
capable students in similar states.

In our experiments, we have shown that the CHF model is able to predict what
capable students would do better than other predictive schemes, especially on an open-ended
programming dataset with high strategic variability. We also showed that the CHF reproduces
human tutor hints about as well as existing hint policies on a simple UML diagram task.
These results indicate that the averaging approach of the CHF is beneficial for prediction, but
that this advantage is not necessarily reflected in higher hint quality, at least for a simple
learning task with low strategic variability.

We note that the Continuous Hint Factory still has several limitations. In particular, the CHF
can only be applied if an edit distance is available which is efficient, takes syntax and
semantics into account appropriately, and yields edits that are viable as next-step-hints for the
student. This is not an issue for the domain of computer programming, as edit distances appear as
a natural fit for syntax-tree-based representations of programs but may be an issue for other
domains. Further, as any data-driven hint approach, hint quality will suffer if the strategy of a
new student is substantially different from anything that the system has seen before.
With regards to evaluation, our assessment of hint quality is not definitive, and it appears likely
that our proposed approach only yields significant advantages compared to existing work on more 
complicated tasks compared to the ones we investigated. Further, we do not yet know how a difference 
in hint quality translates to learning outcomes in students. After all, better hints from the view 
of a tutor may not always yield better learning outcomes, due to difficulties in sense-making or 
lack of prior knowledge on the student's side \cite{Aleven2016}. Finally, we acknowledge that our 
evaluation is rather narrow, including only two learning tasks from different domains.

With regards to future work, it appears promising to integrate the CHF even better with prior
work presented in the literature. In particular, we could take syntactic and unit test 
information into account \cite{Rivers2014}, combine multiple edits instead of single edits 
\cite{Rivers2015}, or apply more sophisticated edit distances as suggested by \citeN{Mokbel2013EDM},
\citeN{Paassen2016Neurocomputing}, as well as \citeN{Price2017EDM}.
Finally, it would be interesting to evaluate the CHF on more learning tasks, especially more 
open-ended learning tasks where the advantages of the CHF are more likely to be visible.

\section{Acknowledgements}

This research was funded by the German Research Foundation (DFG) as part of the project
\enquote{Learning Dynamic Feedback for Intelligent Tutoring Systems} under the grant number HA
2719/6-2 as well as the Cluster of Excellence Cognitive Interaction Technology 'CITEC' (EXC 277),
Bielefeld University and the NSF under grant number \#1432156 \enquote{Educational Data Mining for
Individualized Instruction in STEM Learning Environments} with Min Chi \& Tiffany Barnes as Co-PIs.
We also wish to express our gratitude to our anonymous reviewers and our editors who helped to 
increase the quality of our contribution substantially.